\documentclass{article}
\PassOptionsToPackage{numbers}{natbib}

\usepackage[preprint]{main}

\usepackage[utf8]{inputenc} %
\usepackage[T1]{fontenc}    %
\usepackage{hyperref}       %
\usepackage{url}            %
\usepackage{booktabs}       %
\usepackage{amsfonts}       %
\usepackage{nicefrac}       %
\usepackage{microtype}      %
\usepackage{xcolor}         %

\usepackage{amsmath,amsfonts,bm}

\def\eqref#1{equation~\ref{#1}}

\def\1{\bm{1}}

\DeclareMathAlphabet{\mathsfit}{\encodingdefault}{\sfdefault}{m}{sl}
\SetMathAlphabet{\mathsfit}{bold}{\encodingdefault}{\sfdefault}{bx}{n}

\usepackage[most]{tcolorbox}
\usepackage{helvet}
\usepackage{caption} %
\usepackage{enumitem}
\tcbuselibrary{skins}     
\tcbuselibrary{fitting}
\tcbuselibrary{breakable}

\newtcolorbox{reasonbox}[1][]{%
  enhanced,
  colback=white,
  colframe=blue!20!white,
  boxrule=1.0pt,
  arc=2mm,
  left=1mm,
  right=1mm,
  top=1mm,
  bottom=1mm,
  fonttitle=\normalfont\normalsize,
  fontupper=\sffamily\small,
  title filled,
  colbacktitle=blue!20!white,
  coltitle=black,
  title style={
    draw=black,
    line width=0.5pt,
    rounded corners=1mm,
  },
  #1
}

\usepackage{booktabs}      %
\usepackage{longtable}     %
\usepackage{xcolor,colortbl}  %
\usepackage{adjustbox}     %
\usepackage{array}         %
\usepackage{multirow}      %

\definecolor{highseverity}{RGB}{255,200,200}    %
\definecolor{mediumseverity}{RGB}{255,235,200}  %
\definecolor{lowseverity}{RGB}{200,255,200}     %
\definecolor{headercolor}{RGB}{230,230,230}     %

\tcbset{
    codebox/.style={
        colback=gray!5,
        colframe=gray!50,
        boxrule=0.5pt,
        left=2pt,
        right=2pt,
        top=2pt,
        bottom=2pt,
        fontupper=\ttfamily\small,
        breakable
    }
}

\usepackage{longtable}  
\usepackage{array}
\usepackage{adjustbox}
\usepackage{threeparttable}
\usepackage{xltabular}
\usepackage{enumitem}

\usepackage{booktabs}      %
\usepackage{multirow}      %
\usepackage{xcolor,colortbl}  %
\usepackage{adjustbox}     %

\definecolor{categorygray}{RGB}{230,230,230}
\definecolor{ourpurple}{RGB}{240,235,255}
\definecolor{ourpurple2}{RGB}{220,220,220}  %

\newcommand{\oursrow}{\rowcolor{ourpurple}}

\definecolor{bluetext}{RGB}{0,0,255}
\definecolor{orangetext}{RGB}{255,140,0}
\definecolor{redtext}{RGB}{255,0,0}
\definecolor{tealtext}{RGB}{0,128,128}

\usepackage[utf8]{inputenc}
\usepackage{xcolor}
\usepackage{array}
\usepackage{booktabs}

\definecolor{bluetext}{RGB}{0,0,255}
\definecolor{orangetext}{RGB}{255,140,0}
\definecolor{redtext}{RGB}{255,0,0}
\definecolor{tealtext}{RGB}{0,128,128}

\usepackage{amsthm, amsmath, amssymb, thmtools}

\theoremstyle{plain} %

\theoremstyle{definition} %

\usepackage[ruled,vlined]{algorithm2e}

\usepackage{fontawesome5} %

\usepackage{tabularx} %
\usepackage{multirow}
\usepackage{makecell}
\usepackage{wrapfig}
\usepackage{relsize}
\usepackage[utf8]{inputenc} 
\usepackage{subcaption}
\usepackage{graphicx}
\usepackage{xcolor}
\usepackage{url}
\usepackage{amsmath}
\usepackage{booktabs}     %
\usepackage{multirow}     %
\usepackage{multicol}     %
\usepackage{array}        %
\usepackage{float}
\usepackage{amssymb}
\usepackage{bbm}
\usepackage{bm}
\usepackage{amsmath}
\usepackage{amsthm}
\usepackage{amssymb}
\usepackage{adjustbox}    %
\usepackage{graphicx}     %

\usepackage{threeparttable} %

\usepackage[bottom]{footmisc}
\usepackage{cleveref}
\AddToHook{cmd/appendix/before}{\crefalias{section}{appendix}}

\newcommand{\gettailrelations}{\ensuremath{\mathtt{get\_tail\_relations}}}
\newcommand{\getheadrelations}{\ensuremath{\mathtt{get\_head\_relations}}}
\newcommand{\gettailentities}{\ensuremath{\mathtt{get\_tail\_entities}}}
\newcommand{\getheadentities}{\ensuremath{\mathtt{get\_head\_entities}}}

\newcolumntype{L}[1]{>{\raggedright\arraybackslash}p{#1}}
\newcommand{\lt}{\textless}
\newcommand{\gt}{\textgreater}

\newcommand{\ActSet}{\ensuremath{\mathcal{U}_{\text{ret}}}}

\newcommand{\answer}{\operatorname{Answer}}

\newenvironment{blueblock}{\color{blue}}{\color{black}}

\usepackage{xcolor,pifont}
\newcommand{\dataset}[1]{\texttt{#1}}

\newcommand{\framework}{\textsc{KG-R1}}

\title{Efficient and Transferable Agentic Knowledge Graph RAG via Reinforcement Learning}

\author{
    Junhong Lin$^{1}$\thanks{Equal contribution.}
    \quad
    Shicheng Liu$^{1*}$
    \quad
    Jinyeop Song$^{1}$
    \quad
    Song Wang$^{2}$
    \\
    \textbf{ Julian Shun$^{1}$}
    \quad
    \textbf{Yada Zhu$^{3}$}
    \\[0.35em]
    $^{1}$MIT CSAIL \quad
    $^{2}$University of Virginia \quad
    $^{3}$IBM Research
    \\[0.25em]
    \texttt{\small \{junhong,liushch, yeopjin, jshun\}@mit.edu \quad sw3wv@virginia.edu}\\
    \texttt{\small yzhu@us.ibm.com}
}

\begin{document}

\maketitle

\begin{abstract}
Knowledge-graph retrieval-augmented generation (KG-RAG) couples large language models (LLMs) with structured, verifiable knowledge graphs (KGs) to reduce hallucination and provide reasoning traces. However, current KG-RAG systems often rely on fixed pipelines of multiple LLM modules (e.g., planning, reasoning, and responding), which inflate inference costs and tie performance to specific graph schemas. To address this, we introduce \framework{}, an agentic framework that optimizes KG-RAG through reinforcement learning (RL). Unlike modular workflows, \framework{} uses a single agent that interacts with KGs as its environment, learning to retrieve information at each step and incorporating it into its reasoning and generation in a unified process. Across Knowledge-Graph Question Answering (KGQA) benchmarks, \framework{} demonstrates both \emph{efficiency} and \emph{transferability}---using Qwen 2.5-3B, \framework{} improves answer accuracy with fewer generation tokens than prior multi-module workflow methods that use much larger foundation or fine-tuned models. Furthermore, \framework{} exhibits strong \emph{plug-and-play} capability: after training, maintaining accuracy on unseen KGs without retraining. These properties make \framework{} a promising KG-RAG framework for real-world deployment. Our code is publicly available at \href{https://github.com/junhongmit/KG-R1}{\faGithub~\texttt{github.com/junhongmit/KG-R1/}}.
\end{abstract}

\section{Introduction} 

\renewcommand{\thefootnote}{\fnsymbol{footnote}}
\addtocounter{footnote}{-1}

Retrieval-augmented generation (RAG)~\cite{gao2023retrieval} has gained popularity as a way to enhance large language models (LLMs) with access to external knowledge, thereby reducing hallucination and improving accuracy in knowledge-intensive tasks. Recent research has extended this idea to knowledge graph retrieval-augmented generation (KG-RAG), where \emph{knowledge graphs (KGs)} are leveraged as the retrieval source. A KG is a structured representation of knowledge in the form of entities (nodes) and their relationships (edges) that encodes factual knowledge in a graph format.  Augmenting LLMs with KGs has proven effective not only in mitigating the knowledge bottleneck but also in improving reasoning performance over complex multi-hop relations and enhancing adaptation to continually evolving real-world information~\citep{Sun2024ToG, Luo2024RoG, Wang2025ReKnoS}. These properties make KG-RAG especially promising in high-stakes domains, such as medical consultation and legal analysis~\citep{zhao2025medrag, cui2024chatlawmultiagentcollaborativelegal}.

\begin{figure}[t!]
    % \vspace{-2em}
\centering\includegraphics[width=1.0\textwidth]{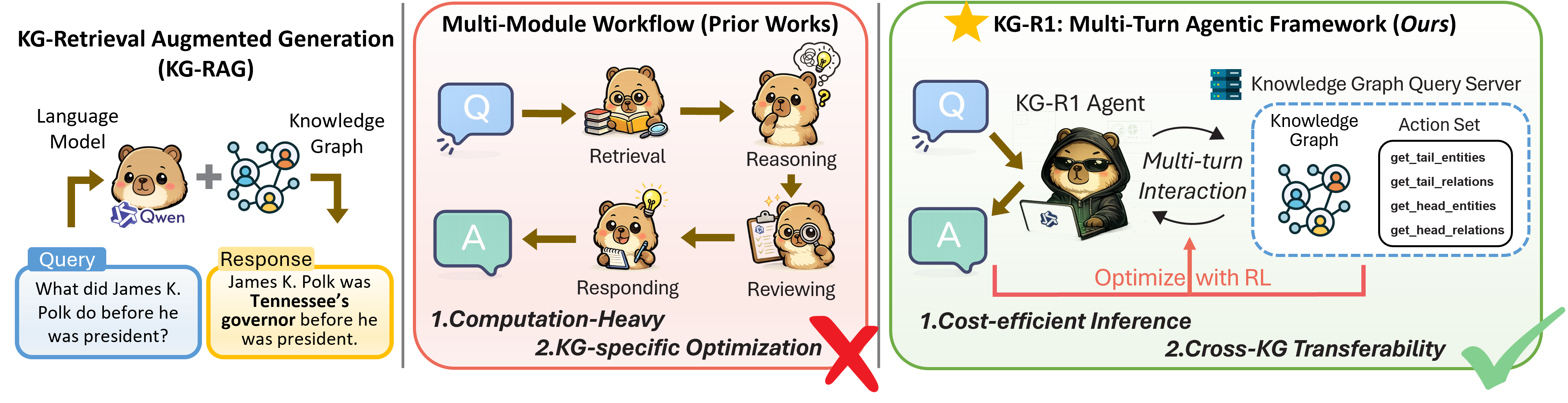}
    % \vspace{-1.5em}
    \caption{Overview of \framework{}, a multi-turn agentic KG-RAG framework. The framework enables cost-efficient inference and demonstrates strong cross-KG transferability.}
    \vspace{-1.5em}
    \label{fig1:overview}
\end{figure}

As shown in \Cref{fig1:overview}, a typical KG-RAG adopts a \emph{modular workflow} consisting of four subtasks: \emph{retrieval} to query facts from KGs, \emph{reasoning} to process retrieved information, \emph{reviewing} to verify logical consistency, and \emph{responding} to synthesize the answer~\citep{baek2023, Sun2024ToG, Luo2024RoG, Wang2025ReKnoS}. Each subtask is handled by specialized LLM modules through two main methods: (i) prompt-based modules with task-specific instructions, often including in-context demonstrations~\citep{baek2023}; and (ii) fine-tuned modules tailored to specific tasks (e.g., SPARQL generation~\citep{cotsparql2024,lee2025sparkle} or relation extraction~\citep{yao2019kgbert}) on specific KGs.

\begin{wrapfigure}{r}{0.45\textwidth}
    \centering
    \vspace{-1.5em}
    \includegraphics[width=\linewidth]{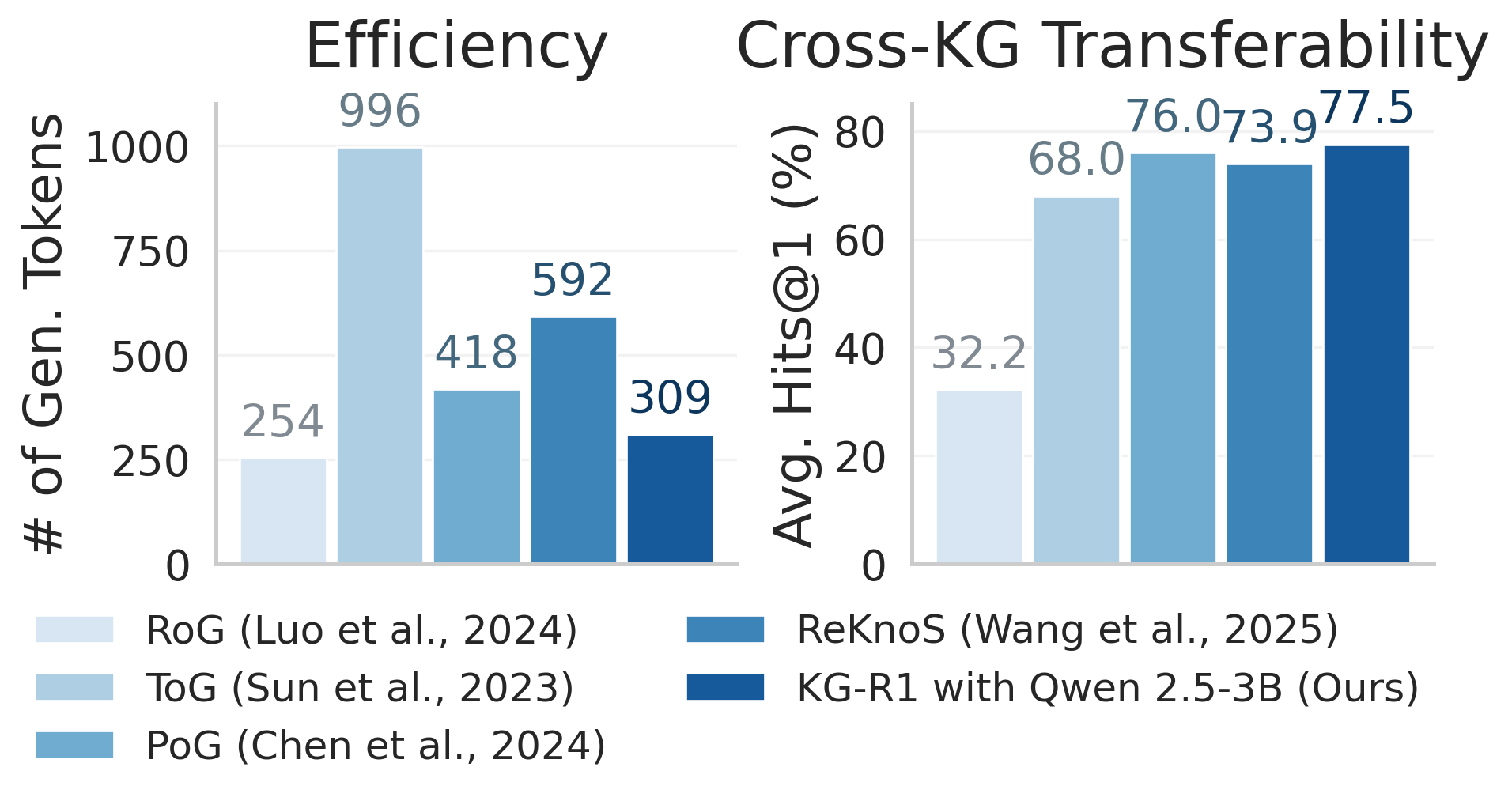}
    \vspace{-1.5em}
    \caption{Prior multi-module methods are costly and do not transfer well across KGs. \emph{Left:} mean end-to-end generated tokens per query on \dataset{WebQSP}~\citep{Yih2016WebQSP}. \emph{Right:} average Hits@1 over five out-of-training KGQA datasets (see \Cref{sec:exp2}). \framework{} achieves \emph{both} low token cost and strong cross-KG transferability.}
\label{fig2}
\vspace{-1.25em}
\end{wrapfigure}

Despite improved reasoning accuracy, the real-world deployment of such workflows faces two key challenges: achieving high reasoning \textbf{effectiveness} while maintaining \textbf{efficiency}, and \textbf{generalization to new or updated KGs}. 
First, prompt-based methods that repeatedly call large foundation LLMs accumulate inference across stages and drive up latency, token usage, and compute (e.g., ToG \citep{Sun2024ToG} and ReKnoS \citep{Wang2025ReKnoS}; see left panel of \Cref{fig2}). Second, these prompted or fine-tuned modules are typically tuned to a particular KG’s domain and schema (entity types, relations, constraints), often via curated in-context examples or KG-specific fine-tuning. As a result, performance does not transfer reliably when the domain shifts, the schema changes, or the system is deployed on a new KG (e.g., RoG \citep{Luo2024RoG}; see right panel of \Cref{fig2}). 

To tackle these challenges, we introduce \textbf{\framework{}}, an agentic KG-RAG system that employs end-to-end multi-turn reinforcement learning~\citep{jin2025searchr1,zeng2025reinforcing, DeepSeek2025R1} to learn effective and transferable KG interaction policies. As shown in \Cref{fig1:overview}, the architecture of \framework{} has two components: a single LLM agent and a KG query server. The KG query server hosts the knowledge graph along with a set of retrieval actions. The LLM agent iteratively performs cycles of short reasoning followed by retrieval actions over multiple turns, with each decision informed by knowledge obtained from the KG retrieval server, and generates a final answer. 
By learning a unified interaction policy, \framework{} avoids redundant multi-module coordination and allocates computation adaptively across reasoning steps, leading to both improved effectiveness and reduced token usage.
\Cref{fig2} demonstrates that \framework{} achieves both high efficiency and strong cross-KG transferability \emph{simultaneously} using a 3B-parameter model, outperforming prior methods.
Our key contributions are summarized as follows:

\noindent \textbf{1. \framework{} Framework.} We introduce an agentic KG-RAG system (Section~\ref{sec:kg_r1}) that replaces multi-module pipelines with a \emph{single} agent for KG-RAG, running on a lightweight KG server. The agent alternates between reasoning and retrieval actions over multiple turns, with the end-to-end trajectory optimized by RL using both \emph{turn-wise} and \emph{outcome-based} reward signals. Turn-wise rewards evaluate individual action effectiveness and adherence to formatting, while global rewards measure answer quality and retrieval relevance.

\noindent \textbf{2. Effective and Efficient Inference.} By consolidating reasoning and retrieval into a \emph{single-agent, near-continuous} workflow, \framework{} achieves competitive reasoning accuracy with a small-parameter model while reducing token usage. This lowers latency and computational cost, making deployment feasible under tight budgets. Experiments demonstrate improved performance and efficiency compared to traditional multi-module workflows (Section~\ref{sec:exp1}). 

\noindent \textbf{3. Plug-and-play Transferability across Diverse KGs.} \framework{} can be easily transferred to diverse KGs and maintains strong KG-RAG performance (Section~\ref{sec:exp2}). The trained \framework{} agent generalizes to new KGs without retraining: backend KGs can be swapped without changing prompts, hyperparameters, or fine-tuning. This generalizability enables zero-shot transfer to unseen KGs.

\section{Related Work}

\noindent\textbf{Knowledge Graph Retrieval-Augmented Generation (KG-RAG).}
KG-RAG augments LLMs with structured KGs to improve quality and compositional reasoning. Early work grounds LLMs' generation by translating natural-language questions into executable graph queries (e.g., SPARQL/Cypher), retrieving relevant subgraphs or answers, and feeding them back to the model~\citep{ouyang2022traininglanguagemodelsfollow, lee2025sparkle}. More recent approaches adopt a modular LLM pipeline over KGs, interleaving natural language reasoning with multi-stage planning, path search, and execution, where each stage uses a prompted or fine-tuned LLM~\citep{Luo2024RoG, Sun2024ToG, Chen2024Pog, Wang2025ReKnoS}. Despite these advances, most systems rely on hand-engineered modules or prompt heuristics tied to a specific KG schema, which induce cost inefficiency and limit generalization. These challenges motivate our \framework{} framework: a single-agent KG-RAG approach that improves efficiency and transfers well to new KGs.

\noindent\textbf{Multi-turn RL for LLMs.}
Reinforcement learning (RL) has become central to equipping LLMs with step-by-step (chain-of-thought) reasoning behavior~\citep{OpenAI2024o1, DeepSeek2025R1}. RL-enhanced models yield substantial gains in math and coding~\citep{le2022coderl,chervonyi2025goldmedal}, and broader complex reasoning tasks.
More recently, RL has been applied to agentic LLMs that invoke external tools (e.g., bash terminals and APIs) or interact with knowledge bases, improving tool use~\citep{Qin2023ToolLLM} and retrieval-augmented generation (RAG)~\citep{jin2025searchr1,wang2025otc} to facilitate tool use or RAG. Building on these advances, our work, \framework{} adopts end-to-end RL as the core optimization algorithm for an agentic KG-RAG framework.

\section{\framework{}: An Agentic KG-RAG Framework}
\label{sec:kg_r1}

\subsection{Problem Definition}

KG-RAG spans many applications, including conversational assistants~\citep{chaudhuri2021grounding}, recommendation systems~\citep{wang2025knowledge}, and open-domain QA~\citep{zhu2025knowledgegraphguidedretrievalaugmented}. In this work, we instantiate and evaluate our approach on Knowledge Graph Question Answering (KGQA), which provides a grounded testbed for KG-RAG: ground-truth answers are tied to a fixed KG, evaluation is verifiable, and intermediate graph reasoning is interpretable. We now formalize knowledge graphs and KGQA tasks.

\noindent\textbf{Knowledge Graphs.} A knowledge graph (KG) is a graph-structured representation of real-world knowledge that encodes factual information as triples of entities and their relationships. We denote a KG as
\(
G=\{\,\langle e,r,e'\rangle \mid e,e'\!\in\!\mathcal{E},\ r\!\in\!\mathcal{R}\,\},
\)
where \(\mathcal{E}\) and \(\mathcal{R}\) denote the sets of entities and relations, respectively, and each triple \(\langle e,r,e'\rangle\) represents a directed edge from entity \(e\) to entity \(e'\) via relation \(r\). For example, there is an edge \texttt{\small{capital\_of}} from entity \texttt{\small{Springfield}} to entity \texttt{\small{Illinois}}.

\noindent\textbf{Knowledge Graph Question Answering (KGQA)} is a reasoning task based on KGs.
Consider a dataset \(D=\{(q, G, A_q)\}\), where \(q\) is a natural language question, \(G\) is a KG, and \(A_q\subseteq\mathcal{E}\) is the ground-truth answer set paired with the question. For each ground truth answer \(e^* \in A_q\), there exists one or more \emph{reasoning paths} in \(G\) that connect anchor entities mentioned in \(q\) to \(e^*\). A \emph{reasoning path} is a sequence of labeled edges \(r_1,\dots,r_\ell\in\mathcal{R}\) instantiated by entities \(e_0,\dots,e_\ell\) such that 
\(Z:\ e_0 \xrightarrow{r_1} e_1 \xrightarrow{r_2} \cdots \xrightarrow{r_\ell} e_\ell,
\)
where \(e_0\) is an anchor entity mentioned in \(q\) and \(e_\ell = e^* \in A_q\), and \(\mathcal{Z}(q, G, A_q)\) denotes the set of all valid reasoning paths for question \(q\) over \(G\) with respect to ground-truth answers \(A_q\). In solving KGQA, given \(q\) and \(G\), a model attempts to discover a subset of valid reasoning paths from \(\mathcal{Z}(q, G, A_q)\) and predicts \(\hat{A}_q\) based on the terminal entities of the discovered paths. The model's performance is evaluated by comparing \(\hat{A}_q\) with the ground truth \(A_q\). As an example, to answer the question ``What is the capital of the U.S. state whose largest city is Chicago?'', a valid reasoning path is\[\text{Chicago}\xrightarrow{\texttt{located\_in\_state}}\text{Illinois}\xrightarrow{\texttt{capital\_of}}\text{Springfield}.\]
This path belongs to \(\mathcal{Z}(q, G, A_q)\) and leads to the correct prediction \(\hat{A}_q=\{\text{Springfield}\}\).

\subsection{\framework{} Framework}

\begin{figure}[t]
    \centering
    \vspace{-2em}
    \includegraphics[width=\textwidth]{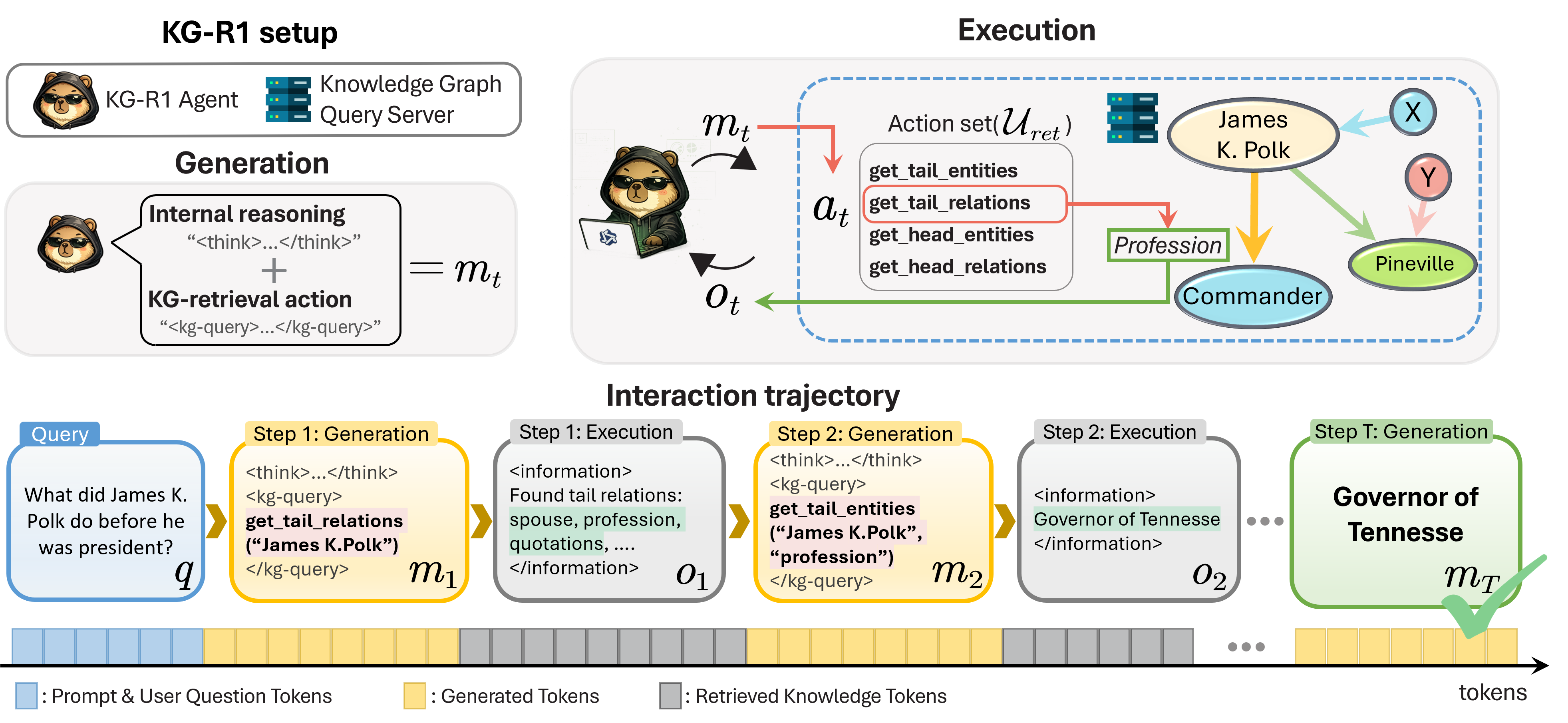}
    \vspace{-1em}
    \caption{\framework{} framework: a single LLM agent undergoes a multi-turn generation-execution loop with a schema-agnostic KG retrieval server and responds with the final answer. }
    \label{fig3}
% \vspace{-2em}
\end{figure}

\framework{} casts KG-RAG as a multi-turn interaction with a KG interface (KG query server). We prioritize two design principles. First, we design a \emph{single-agent} architecture that simplifies deployment and enables efficient, low-cost inference. Second, we build a \emph{schema-agnostic} KG retrieval server that avoids KG-specific assumptions and remains portable across different KGs.
Given a KGQA dataset \(D=\{(q,G,A_q)\}\), we set up a KG retrieval server and a \framework{} agent. 

\noindent\textbf{KG Query Server.}
\label{subsub:kg-server}
The server hosts the knowledge graph \(G\) and provides a set of retrieval actions \(a \in \ActSet\) that enable graph traversal through 1-hop operations:
\[
\mathcal{A}_{\mathrm{ret}}
= \bigl\{ \scalebox{0.9}{$\gettailrelations,\ \getheadrelations,\ 
\gettailentities,\ \getheadentities$} \bigr\}
\]

\vspace{-2.5em}
\begin{align*}
\mathtt{get\_tail\_relations} (e)
&:= \{\, r \in \mathcal{R} \mid \exists e'\in\mathcal{E}:\ (e,r,e')\in G \,\},\\
\mathtt{get\_head\_relations} (e')
&:= \{\, r \in \mathcal{R} \mid \exists e\in\mathcal{E}:\ (e,r,e')\in G \,\},\\
\mathtt{get\_tail\_entities} (e,r)
&:= \{\, e' \in \mathcal{E} \mid (e,r,e')\in G \,\},\\
\mathtt{get\_head\_entities} (r,e')
&:= \{\, e \in \mathcal{E} \mid (e,r,e')\in G \,\}.
\end{align*}
\vspace{-1.5em}

The retrieval action set \(\ActSet\) is sufficient for \emph{realizing} any reasoning path \(Z:\ e_0 \xrightarrow{r_1} e_1 \xrightarrow{r_2} \cdots \xrightarrow{r_\ell} e_\ell\) as a finite action sequence that arrives at \(e_\ell \in A_q\) (i.e., the terminal entity is in the answer set). Forward traversal along \(Z\) is implemented by \(\mathtt{get\_tail\_entities} (e_{i-1},r_i)\) for \(i=1,\dots,\ell\) and backward traversal is implemented by \(\mathtt{get\_head\_entities} (r_i,e_i)\). 
This action design \emph{guarantees} completeness of path realization, as formalized in Remark~\ref{prop:integrity}, and provides a schema-agnostic interaction interface that enables transfer across KGs. Full formal statements and additional properties are provided in Appendix~\ref{app:kg-r1-proofs}.

\begin{restatable}[\textbf{Retrieval Action Set Completeness}]{remark}{completeness}
\label{prop:integrity}
For any reasoning path \(Z:\ e_0 \xrightarrow{r_1}\cdots\xrightarrow{r_\ell} e_\ell\) in \(G\), there exists an action sequence in \(\ActSet\) of length at most \(\ell{+}1\) whose output includes \(e_\ell\).
\end{restatable}

\noindent\textbf{\textsc{KG-R1} Agent.} 
We model a single \emph{LLM agent} that interacts with a KG query server in a multi-turn setting. The agent undergoes initialization followed by a loop of generation and execution (see Figure~\ref{fig3}). At initialization, a lightweight base LLM is configured with an instruction prompt \(p\) (see the box below for the prompt template) containing general reasoning instructions, the user question \(q\), and the KG retrieval server instructions (Table~\ref{tab:kg_server_instruction}). At each turn \(t \le H\), where $H$ is the maximum turn limit, the agent first undergoes the \textbf{generation} phase where it produces a response \(m_t\) comprising two parts: (1) an internal reasoning wrapped in \texttt{<think>...</think>}, and (2) an action, which is either a KG retrieval action wrapped in \texttt{<kg-query>...</kg-query>}, or a final answer wrapped in \texttt{<answer>...</answer>}. 

\begin{tcolorbox}[colback=white, colframe=black, title=Prompt template for \texttt{KG-R1}, boxrule=0.5pt, left=2mm, right=2mm, top=0mm, bottom=0.5mm]
\footnotesize
\label{tab:instruction_prompt}
You are a helpful assistant. Answer the given question. You can query from knowledge base provided to you to answer the question. You can query knowledge up to [H] times. You must first conduct reasoning inside \textcolor{bluetext}{$<$think$>$...$<$/think$>$}. If you need to query knowledge, you can set a query statement between \textcolor{orangetext}{$<$kg-query$>$...$<$/kg-query$>$} to query from knowledge base after \textcolor{bluetext}{$<$think$>$...$<$/think$>$}. When you have the final answer, you can output the answer inside \textcolor{redtext}{$<$answer$>$...$<$/answer$>$}. \\
KG Query Server Instruction : [{KG\_query\_server\_instruction}] \\
Question: [question]. \\
Assistant:
\end{tcolorbox}

In the following \textbf{execution} phase, we parse \(m_t\) into an \emph{action} \(a_t \in \ActSet \cup \{\answer\}\) using an exact-match parser \(\Psi\) (i.e., \(a_t=\Psi(m_t)\)). If \(a_t \in \ActSet\), executing it yields an \emph{observation} \(o_{t+1}\) (i.e., retrieved entities, relations, or an error message in case the action does not generate a valid retrieval), which is appended to the dialogue context prior to the next turn. If \(a_t=\answer\), the interaction terminates: the content inside \texttt{<answer>...</answer>} is extracted and post-processed (normalization, deduplication, and entity resolution) to produce the predicted answer set \(\hat{A}_q\). Given an interaction trajectory $\tau = (p,\,(m_1,o_1),\ldots,(m_t,o_{t}))$ with $t \leq H$, the \framework{} agent’s \emph{policy} \(\pi_\theta(m_t \mid \text{context}_t)\) is defined over token sequences \(m_t\) and governs how textual responses are generated. 

\subsection{\framework{} Training with RL}
\label{sub:kg_r1_rl}

Our goal is to learn a policy \(\pi_\theta(m_t \mid \text{context}_t)\) for the \framework{} agent that interacts with the knowledge graph $G$ over multiple turns to incrementally retrieve informative reasoning paths and ultimately produce the correct answer \(\hat{A}_q =  A_q\). We optimize this policy using reinforcement learning with a GRPO-style objective \citep{DeepSeek2025R1}, following recent advances in multi-turn agent optimization~\citep{qian2025toolrl, jin2025searchr1, zeng2025reinforcing}. An overview of the training procedure is provided in Algorithm~\ref{alg:kg_r1_training} in the \Cref{sec:kgr1_details}. %

\noindent\textbf{Reward Objective.} 
\label{subsub:Reward_Function} To effectively train the \framework{} agent, we combine verifiable \emph{turn} rewards with outcome-level \emph{global} rewards. \textbf{Turn rewards} (\(r^{\text{turn}}_{t}\)) provide local signals at each turn as the sum of three components: (i) \emph{format validity} \(v_{\mathrm{fmt}}(m_t)\) checks that \(m_t\) contains both reasoning and a well-formed action \(a_t\in\mathcal{A}\);
(ii) \emph{KG query} \(v_{\mathrm{kg}}(a_t,o_{t+1})\) checks that executing \(a_t\) yields meaningful, schema-valid retrieval in \(o_{t+1}\); and
(iii) \emph{answer consistency} \(v_{\mathrm{ans}}(m_T)\) checks the final answer’s format/consistency on the final turn. The turn rewards are computed as follows with weights $w_{fmt}$, $w_{kg}$, and $w_{ans}$: 
\begin{equation}
r^{\mathrm{turn}}_{t}
= w_{\mathrm{fmt}}\,v_{\mathrm{fmt}}(m_t)
+ w_{\mathrm{kg}}\,v_{\mathrm{kg}}(a_t,o_{t+1})
+ w_{\mathrm{ans}}\,v_{\mathrm{ans}}(m_T).
\end{equation}

\textbf{Global rewards} summarize trajectory-level outcomes as the sum of the following:
(i) a binary answer correctness signal $\mathrm{H1}$ that is $1$ if the predicted answer \(\hat{A}_q\) is included in the ground-truth answer set \(A_q\), and $0$ otherwise; and
(ii) a \emph{retrieval score} $v_{ret}$ that is \(1\) if any ground-truth entity appears anywhere in the retrieved information along the executed reasoning path, and \(0\) otherwise. 
\begin{equation}
R^{\mathrm{global}}
= w_{\mathrm{H1}}\cdot \mathrm{H1}\!\bigl(\hat{A}_q, A_q\bigr)
\;+\;
w_{\mathrm{ret}}\cdot v_{ret}
\end{equation}
\noindent\textbf{Group-relative Turn-level Credit Assignment and Optimization.} \label{subsub:rl_algo}
To assign credit across multi-turn trajectories, we adopt a group-relative, turn-level credit assignment strategy \cite{zeng2025reinforcing}. 
We collect $N$ rollouts per query $q$. For each trajectory $\tau^{(n)}$, we compute turn rewards $r^{\mathrm{turn},(n)}_t$ and a global reward $R^{\mathrm{global},(n)}$, and combine them into a per-turn return with $\lambda$ as the normalization factor:
\begin{equation}
G_t^{(n)} \;=\; r^{\mathrm{turn},(n)}_t \;+\; \lambda\,R^{\mathrm{global},(n)}.
\end{equation}
Let $\mathcal{S} \!=\! \{(n,t)\!:\! t \le T^{(n)}\}$ be the set of all turns across the $N$ rollouts. The turn-level advantage $A_t^{n}$ is calculated using a \emph{single} group baseline $\bar{G}$ that averages over $\mathcal{S}$:
\[
A_t^{(n)}=\frac{G_t^{(n)}-\bar G}{\sigma_G+\varepsilon}
\quad\text{where}\quad
\bar G=\tfrac{1}{|\mathcal S|}\!\sum_{(n,t)\in\mathcal S} G_t^{(n)},\ 
\sigma_G=\sqrt{\tfrac{1}{|\mathcal S|}\!\sum_{(n,t)\in\mathcal S}\!\big(G_t^{(n)}-\bar G\big)^2},
\]
where $\varepsilon$ is a constant for numerical stability. Intuitively, this measures each turn's contribution relative to the group average, reinforcing above-average decisions and discouraging below-average ones.

\noindent \textbf{RL update.} Using the turn-level advantage $A_t^{(n)}$, we optimize the agent's policy $\pi_\theta$ with a GRPO-style clipped objective $\mathcal{J}$:
\[
\mathcal{J}(\theta)=\mathbb{E}\!\Bigg[\sum_{n,t,i}
\Big(
\min\!\Big(\rho_{t,i}^{(n)}\widetilde{A}_{t,i}^{(n)},\ \operatorname{clip}\big(\rho_{t,i}^{(n)},1-\epsilon,1+\epsilon\big)\,\widetilde{A}_{t,i}^{(n)}\Big)
-\beta\mathrm{KL}\!\big(\pi_\theta(\cdot\mid h_{t,i}^{(n)})\|\pi_{0}(\cdot\mid h_{t,i}^{(n)})\big)
\Big)
\Bigg],
\]
where $\widetilde{A}_{t,i}^{(n)}=m_{t,i}^{(n)}A_t^{(n)}$ applies the same turn-level advantage to all tokens within a turn via a mask $m_{t,i}^{(n)}$, $\operatorname{clip}$ bounds updates to improve stability, $n$, $t$, and $i$ index trajectories, turns, and tokens, respectively. 
The current policy is $\pi_\theta$, the behavior policy $\pi_{\theta_{\text{old}}}$ is used for sampling, and $\pi_0$ is a fixed reference policy used only for KL regularization (weight $\beta$). The importance ratio
$\rho_{t,i}^{(n)}=\exp\!\big(\log \pi_\theta(y_{t,i}^{(n)}\!\mid h_{t,i}^{(n)})-\log \pi_{\theta_{\text{old}}}(y_{t,i}^{(n)}\!\mid h_{t,i}^{(n)})\big)$ corrects off-policy sampling at each token $i$. We maximize $\mathcal{J}(\theta)$ via gradient ascent. The proposed combination of structured rewards and group-relative credit assignment yields stable and informative learning signals for multi-turn decision-making, leading to improved reasoning quality and final answer correctness. We provide comprehensive ablation studies of each component of our reward design in Section~\ref{sec:exp3}.

\subsection{Cross-KG Transfer: Plug-and-Play}~\label{sec:plug_and_play}
\vspace{-.1in}

We propose a plug-and-play approach for \framework{} that enables KG-RAG to operate across different knowledge graphs without retraining. Let \(\pi_{\theta \mid \mathcal{D}}\) denote the policy of the \framework{} agent trained on a KGQA dataset \(\mathcal{D}=\{(q,G,A_q)\}\). For a new KGQA dataset \(\mathcal{D}^*=\{(q^*,G^*,A_{q^*})\}\), we replace the backend KG of the retrieval server with \(G^*\) (\emph{plug}) and evaluate \(\pi_{\theta \mid \mathcal{D}}\) on \(\mathcal{D}^*\) without any modification (\emph{play}). This approach requires no task-specific fine-tuning, data relabeling, or architectural changes. 

This plug-and-play capability stems from two design choices in \framework{}. First, the KG server exposes a schema-agnostic action set consisting solely of 1-hop retrieval operations, which are universally compatible with directed KGs. This contrasts with SPARQL-generation approaches in prior work that depend on KG-specific syntax and schema knowledge. Second, the agent’s strategy for retrieving and using KG evidence is learned in a way that is independent of any particular schema, enabling immediate transfer to new domains.

\section{Experiments}
\label{sec:experiments}

We evaluate \framework{} through two research questions: \textbf{RQ1} (Effectiveness and Efficiency): \emph{Can \framework{} learn an effective and efficient KG-RAG policy when trained on a target KG?}; \textbf{RQ2} (Cross-KG Transferability): \emph{Does \framework{} generalize to unseen KGs without additional training?}
For both research questions, we compare against established LLM-only and LLM+KG baselines to assess effectiveness, efficiency, and transferability.

\subsection{Experimental Setup}

\noindent \textbf{Models.}
We use Qwen2.5-3B-it~\citep{qwen2025qwen25technicalreport} as our base model. For the other baseline methods, we use Qwen2.5-7B-it, Qwen2.5-72B-it, Qwen3-235B-A22B-Instruct-2507~\citep{yang2025qwen3}, and Llama2-7B-it (only for the fine-tuned planning in RoG~\citep{Luo2024RoG}), following the setups used in prior work.

\noindent \textbf{Datasets.}
We mainly evaluate \framework{} in-domain on (1) \dataset{WebQSP}~\citep{Yih2016WebQSP}: natural-language questions over Freebase, mostly 1–2 hop questions, using the official 3{,}098/1{,}639 train/test QA split; and (2) \dataset{ComplexWebQuestions (CWQ)}~\citep{Talmor2018CWQ}: compositional, multi-hop questions over Freebase with a 24{,}649/3{,}531 train/test QA split. For scalability, we extract 2-hop subgraphs for each question as in RoG~\citep{Luo2024RoG}. See Appendix~\ref{appendix:datasets} for detailed sources.
We evaluate the plug-and-play approach on diverse KGQA benchmarks spanning two categories of KGs: (1) \emph{Freebase} type---\dataset{SimpleQA}~\citep{bordes2015simplequestions} and \dataset{GrailQA}~\citep{gu2021grailqa}, which share the Freebase KG but differ in question complexity and reasoning path distributions; and (2) \emph{Wikidata} type---\dataset{T-REx}~\citep{elsahar2018trex} and \dataset{QALD-10en}~\citep{usbeck2023qald10} for cross-schema generalization. Appendix~\ref{appendix:datasets} provides a summary of dataset sources, schema, and evaluation splits.

\noindent \textbf{Metrics.} Following prior works~\citep{Luo2024RoG, Sun2024ToG, Chen2024Pog, park2026prograg} to evaluate KGQA performance, we use \textit{Hits@1} (1 if the ground-truth answer appears in the single search, and 0 otherwise). For efficiency, we report \emph{total tokens} and \emph{generated tokens} (completion-only), measured end-to-end per query, aggregated across all turns. 
This serves as a proxy for inference time and compute cost. For fair comparison, all token counts are computed with the Qwen2.5 tokenizer. We also report the \emph{number of modules} to compare workflow complexity, and we analyze single-query latency and batched throughput on a single NVIDIA H100 GPU.

\noindent \textbf{Baselines.}
We compare three classes of approaches: (1) \emph{LLM-only} methods that do not access an external KG: IO~\citep{brown2020gpt3} and Chain-of-Thought (CoT)~\citep{wei2022chain} prompting; (2) \emph{prior KG-augmented} pipelines: \textsc{RoG}~\citep{Luo2024RoG}, which uses a \emph{fine-tuned} modular workflow with LLaMA2-7B-it as the planner and we use the Qwen family as reasoner, and \textsc{ToG}~\citep{Sun2024ToG} and \textsc{ReKnoS}~\citep{Wang2025ReKnoS}, which use a \emph{prompt-based} modular workflow; and (3) our method \framework{}. For LLM-only baselines, we employ LLM-as-Judge~\citep{zheng2023llmasajudge} using gpt-4o-mini~\citep{openai2024gpt4omini} to semantically match predicted answers to ground truth. We report the average and standard deviation over 5 runs for all models and baselines. Full details for the baselines are provided in Appendix~\ref{appendix:baselines}. 

\noindent\textbf{Implementation details.} \label{sec:kg_r1_implementation}
We use VERL~\citep{sheng2024hybridflow} and vLLM-backed decoding~\citep{kwon2023efficient} for implementing the \framework{} agent. Training uses distributed optimization with gradient checkpointing and offloading. Unless stated otherwise, we set \(H=5\) as the maximum number of turns and \(N=16\) rollouts per query. %
Additional implementation details (e.g., hyperparameters) are provided in Appendix~\ref{appendix:implementation_details}.

\begin{wrapfigure}{t}{.5\textwidth}
  \vspace{-4em}
  \centering
  \includegraphics[width=\linewidth]{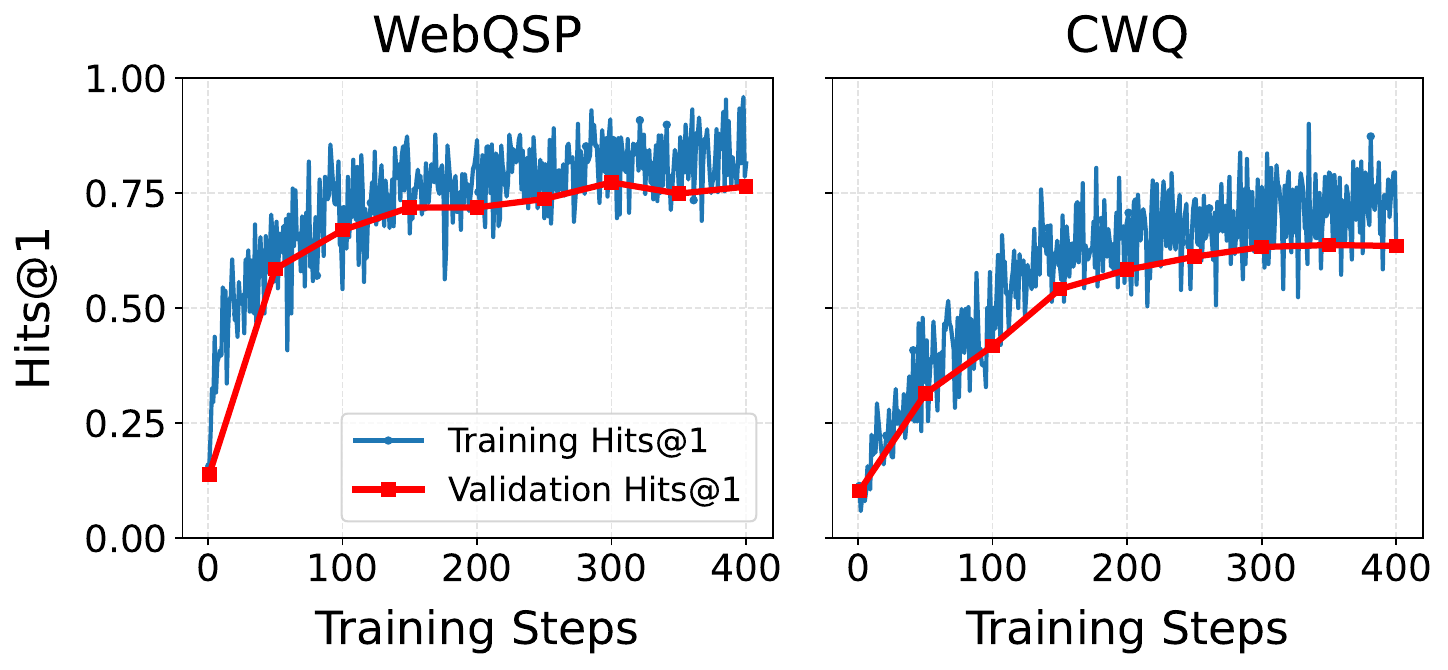}
  \captionof{figure}{Hits@1 over \framework{} training on \dataset{WebQSP} and \dataset{CWQ} for Qwen2.5-3B-it. %
  }
  \label{fig:training_main}
  \vspace{-2em}
\end{wrapfigure}
\subsection{Training Stability and Effectiveness}
Figure~\ref{fig:training_main} shows steady improvements in Hits@1 that gradually plateau as training converges for \framework{} on both \dataset{WebQSP} and \dataset{CWQ}. The validation Hits@1 closely follows the training curve, indicating that the learned policy generalizes well to unseen data in the validation set without significant overfitting. Overall, \framework{} training achieves stable and consistent gains in end-to-end KG-RAG performance.

\subsection{RQ1: Effectiveness and Efficiency Results}
\label{sec:exp1}

\definecolor{first}{RGB}{237, 106, 89}
\definecolor{second}{RGB}{255, 192, 0}
\definecolor{third}{RGB}{91, 155, 213}
\newcommand{\res}[2]{#1\textsubscript{ $\pm$ #2}}
\newcommand{\fir}[2]{\textcolor{first}{\textbf{#1\textsubscript{$\pm$#2}}}}
\newcommand{\sed}[2]{\underline{#1\textsubscript{ $\pm$ #2}}}
\newcommand{\thi}[2]{{#1\textsubscript{ $\pm$ #2}}}
\newcommand{\firi}[1]{\textcolor{first}{\textbf{#1}}}
\newcommand{\seci}[1]{\underline{#1}}
\newcommand{\thii}[1]{{#1}}
\newcommand{\colorres}[3]{\textcolor{#1}{\textbf{#2} \textsubscript{$\pm$ \textbf{#3}}}}

\begin{table*}[!t]
\caption{Performance and efficiency comparison of KGQA methods on WebQSP and CWQ datasets. The maximum turn number is set to $H=5$. All scores are percentages.}
\label{tab:webqsp_cwq_performance_efficiency_comparison_colored}
\vspace{-0.5em}
\begin{center}
\renewcommand{\arraystretch}{1.05}
\small
\resizebox{\textwidth}{!}{%
\begin{tabularx}{1.08\textwidth}{c@{\hspace{0.2em}}c@{\hspace{0.6em}}c@{\hspace{0.2em}}c@{\hspace{0.6em}}c@{\hspace{0.5em}}c@{\hspace{0.5em}}c@{\hspace{0.5em}}c@{\hspace{0.5em}}c}
\toprule
& \multicolumn{2}{c}{\thead{\normalsize Datasets $\rightarrow$}} & \multicolumn{3}{c}{\dataset{WebQSP}} &  \multicolumn{3}{c}{\dataset{CWQ}}
\\
\cmidrule(lr){4-6} \cmidrule(lr){7-9}
& \thead{\normalsize Methods} & \thead{Modules}
& \thead{Hits@1 (\%)} & \thead{Total Tokens} & \thead{Gen Tokens}
& \thead{Hits@1 (\%)} & \thead{Total Tokens} & \thead{Gen Tokens}
\\
\midrule
\multirow{8}{*}{\rotatebox{90}{Qwen 2.5-7B}}
& {IO} & 1
& \res{21.37}{0.07} & \res{151.59}{0.03} & \res{9.29}{0.03}
& \res{16.63}{0.01} & \res{159.76}{0.01} & \res{8.57}{0.01}
\\
& {CoT} & 1
& \res{17.27}{0.09} & \res{260.43}{0.02} & \res{49.13}{0.02}
& \res{14.95}{0.02} & \res{292.46}{0.01} & \res{72.27}{0.01}
\\
& {SC} & 1
& \res{19.26}{0.19} & \res{2098.01}{0.92} & \res{407.60}{0.92}
& \res{17.63}{0.17} & \res{2364.62}{1.24} & \res{603.13}{1.24}
\\
\cmidrule(lr){2-9}
& {RoG} & 2
& \res{79.89}{0.35} & \res{814.61}{0.70} & \res{246.95}{0.67}
& \res{46.32}{0.49} & \res{1219.72}{0.61} & \res{259.46}{0.58}
\\
& {ToG} & 4
& \res{65.00}{4.63} & \res{6455.04}{753.60} & \res{941.52}{109.27}
& \res{32.25}{0.43} & \res{8396.80}{87.06} & \res{1358.54}{17.24}
\\
& {PoG} & 5
& \res{64.81}{0.84} & \res{4474.30}{91.92} & \res{241.14}{5.17}
& \res{32.00}{2.53} & \res{8293.67}{1454.76} & \res{369.84}{45.59}
\\
& {ReknoS} & 3
& \res{61.22}{5.46} & \res{2822.34}{209.87} & \res{558.43}{45.52}
& \res{53.45}{3.03} & \res{2908.55}{222.47} & \res{602.08}{39.33}
\\
\midrule
\multirow{8}{*}{\rotatebox{90}{Qwen 2.5-72B}}
& {IO} & 1
& \res{24.66}{0.07} & \res{153.50}{0.02} & \res{11.20}{0.02}
& \res{22.20}{0.02} & \res{162.04}{0.03} & \res{10.85}{0.03}
\\
& {CoT} & 1
& \res{21.84}{0.04} & \res{271.46}{0.02} & \res{60.16}{0.02}
& \res{22.92}{0.04} & \res{296.81}{0.04} & \res{76.62}{0.04}
\\
& {SC} & 1
& \res{22.62}{0.07} & \res{2177.75}{1.23} & \res{487.34}{1.23}
& \res{24.76}{0.27} & \res{2385.73}{0.51} & \res{624.24}{0.51}
\\
\cmidrule(lr){2-9}
& {RoG} & 2
& \res{81.63}{3.33} & \res{812.32}{7.81} & \res{244.53}{7.79}
& \res{45.05}{2.87} & \res{1205.48}{2.51} & \res{245.34}{2.51}
\\
& {ToG} & 4
& \res{74.47}{0.25} & \res{6810.27}{39.73} & \res{942.91}{5.80}
& \res{43.18}{1.02} & \res{9024.77}{271.68} & \res{1358.32}{41.02}
\\
& {PoG} & 5
& \res{72.88}{6.57} & \res{4759.10}{210.29} & \res{316.25}{16.85}
& \res{43.62}{1.38} & \res{7969.21}{1631.58} & \res{503.29}{116.35}
\\
& {ReknoS} & 3
& \res{75.23}{9.25} & \res{2472.90}{115.94} & \res{488.44}{39.87}
& \res{60.88}{2.54} & \res{2278.71}{196.08} & \res{548.52}{27.79}
\\
\midrule
\multirow{8}{*}{\rotatebox{90}{Qwen 3-235B}}
& {IO} & 1
& \res{24.01}{0.18} & \res{136.05}{0.06} & \res{14.75}{0.06}
& \res{20.22}{0.09} & \res{147.36}{0.16} & \res{17.17}{0.16}
\\
& {CoT} & 1
& \res{26.94}{0.22} & \res{274.66}{0.32} & \res{84.36}{0.32}
& \res{24.38}{0.33} & \res{317.42}{0.31} & \res{118.23}{0.31}
\\
& {SC} & 1
& \res{27.60}{0.15} & \res{2200.20}{1.26} & \res{677.79}{1.26}
& \res{26.13}{0.23} & \res{2551.71}{0.78} & \res{958.22}{0.78}
\\
\cmidrule(lr){2-9}
& {RoG} & 2
& \res{81.68}{0.24} & \res{960.58}{5.90} & \res{254.31}{0.11}
& \res{48.94}{0.39} & \res{1217.51}{10.26} & \res{252.45}{5.18}
\\
& {ToG} & 4
& \res{75.67}{0.37} & \res{6831.66}{28.39} & \res{996.12}{5.21}
& \res{39.82}{0.73} & \res{5389.00}{33.01} & \res{802.97}{9.84}
\\
& {PoG} & 5
& \res{79.13}{1.89} & \res{6818.26}{908.76} & \res{418.26}{49.36}
& \res{44.69}{0.27} & \res{10177.37}{1859.27} & \res{548.77}{82.56}
\\
& {ReknoS} & 3
& \res{84.73}{1.78} & \res{2795.53}{139.20} & \res{592.38}{4.18}
& \res{65.63}{4.86} & \res{2677.92}{281.97} & \res{684.38}{14.06}
\\
\midrule
\oursrow
& \thead{\textbf{\framework{} \scriptsize{(Ours)}}\\\scriptsize{Qwen2.5-3B}} & 1
& \res{82.75}{0.04} & \res{3237.17}{30.34} & \res{308.67}{0.56}
& \res{65.31}{0.14} & \res{3205.95}{42.34} & \res{311.14}{0.52}
\\
\bottomrule
\end{tabularx}
}
\vspace{-1.5em}
\end{center}
\end{table*}

\noindent \textbf{\framework{} demonstrates strong accuracy on trained KGs.}
Table~\ref{tab:webqsp_cwq_performance_efficiency_comparison_colored} reports Hits@1 and token efficiency on \dataset{WebQSP} and \dataset{CWQ}. Despite being trained from a lightweight Qwen2.5-3B base model, \framework{} achieves the strongest overall performance among all compared methods on \dataset{WebQSP}, reaching a Hits@1 of \textbf{82.75}, outperforming RoG built on Qwen2.5-7B (79.89), Qwen2.5-72B (81.63), and even Qwen3-235B (81.68). It also exceeds or is comparable to stronger multi-module baselines such as ToG (75.67), PoG (79.13), and ReKnoS (84.73) while using a substantially smaller backbone.
On the more compositional \dataset{CWQ}, \framework{} achieves a Hits@1 of \textbf{65.31}, improving over RoG by a large margin across all scales (46.32--48.94), and surpassing ReKnoS built on Qwen2.5-7B (53.45), Qwen2.5-72B (60.88), and Qwen3-235B (55.63). Notably, \framework{} remains highly competitive with ReKnoS built on Qwen3-235B (65.31 vs.\ 65.63 on \dataset{CWQ}), despite using a Qwen2.5-3B backbone that is nearly two orders of magnitude smaller. This highlights that reinforcement learning over structured KG actions can substantially improve reasoning effectiveness beyond what is achievable by model scaling alone.

\noindent \textbf{\framework{} achieves efficient inference with reduced generation cost.}
Table~\ref{tab:webqsp_cwq_performance_efficiency_comparison_colored} also reports total and generated tokens. On \dataset{WebQSP}, \framework{} uses only \textbf{308.67} generation tokens, which is 3.05$\times$ fewer than ToG (941.52) and 1.81$\times$ fewer than ReKnoS (592.38), while achieving higher Hits@1 than ToG and comparable performance to ReKnoS. On \dataset{CWQ}, \framework{} generates \textbf{311.14} tokens, reducing generation cost by 2.58$\times$ compared to ToG (802.97) and 2.20$\times$ compared to ReKnoS (684.38), while achieving comparable answer accuracy to ReKnoS and outperforming ToG. Compared with the most token-efficient KG baseline RoG, \framework{} uses moderately more generation tokens (308.67 vs.\ 254.31 on WebQSP; 311.14 vs.\ 252.45 on CWQ), but delivers substantially higher answer accuracy (+1.07 to +20.26 absolute Hits@1 depending on model scale and dataset). This indicates a favorable effectiveness-efficiency tradeoff: \framework{} preserves competitive token efficiency while substantially improving reasoning quality.

\noindent \textbf{Single-agent design reduces system complexity.} Beyond token efficiency, \framework{} simplifies the KG-RAG pipeline into a single-agent architecture. Prior methods typically require 2--5 modules (e.g., retrieval, planning, verification, and answering), whereas \framework{} performs retrieval and reasoning within a unified policy using only one module. This reduces deployment complexity and avoids error propagation across stages. Despite this simplification, \framework{} remains competitive or superior to multi-module systems across both datasets, showing that structured RL can effectively consolidate KG interaction and reasoning into a lightweight unified agent.

\noindent \textbf{Latency and throughput analysis.} 
On a single NVIDIA H100 GPU, we measure (i) single-query end-to-end latency and (ii) batched throughput. The single-query latency averages \(6.4\pm1.5\) s per question, and the batched throughput reaches \(3.7\) samples per second with a batch size of 64. The results suggest the feasibility of \framework{}'s single-node deployment. See Appendix~\ref{appendix:latency_memory_delopyment} for the full results.

\subsection{RQ2: Cross-KG Transferability}
\label{sec:exp2}

\begin{table*}[!t]
\caption{Zero-shot cross-KG transferability of \framework{} (Hits@1, \%). Agents are trained on WebQSP or CWQ and evaluated on new benchmarks by \emph{swapping only} the KG-backend server (no policy retraining). \textbf{AVG} is averaged across QA samples. The maximum turn number is set to $H=5$.}
\label{tab:generalization_kgqa}
\vspace{-0.5em}
\begin{center}
\renewcommand{\arraystretch}{1.05}

\small
\resizebox{0.8\textwidth}{!}{%
\begin{tabularx}{0.94\textwidth}{c@{\hspace{0.7em}}c@{\hspace{0.7em}}ccccc}%
\toprule
& & \multicolumn{2}{c}{Freebase-based} &  \multicolumn{2}{c}{Wikidata-based} \\
\cmidrule(lr){3-4} \cmidrule(lr){5-6}
& \thead{\normalsize Methods$\downarrow$} &\dataset{SimpleQA} & \dataset{GrailQA} & \dataset{T-REx} & \dataset{QALD-10en} & \textbf{AVG}
\\
\midrule
\multirow{8}{*}{\rotatebox{90}{Qwen 2.5-7B}}
& {IO}
& \res{10.30}{0.10} & \res{8.07}{0.06}
& \res{22.79}{0.05} & \res{17.43}{0.01}
& 18.84
\\
& {CoT}
& \res{7.07}{0.06} & \res{7.73}{0.12}
& \res{17.38}{0.04} & \res{21.05}{0.06}
& 14.82
\\
& {SC}
& \res{8.70}{0.20} & \res{8.60}{0.44}
& \res{20.01}{0.12} & \res{20.32}{0.46}
& 16.93
\\
\cmidrule(lr){2-7}
& {RoG}
& \res{12.97}{0.35} & \res{17.67}{0.15}
& \res{22.37}{0.27} & \res{29.33}{0.46}
& 20.76
\\
& {ToG}
& \res{52.37}{0.80} & \res{54.73}{4.62}
& \res{58.55}{1.11} & \res{50.95}{0.92}
& 56.84
\\
& {PoG}
& \res{66.38}{0.39} & \res{69.88}{0.84}
& \res{75.69}{0.08} & \res{46.07}{0.25}
& 72.28
\\
& {ReknoS}
& \res{56.43}{1.22} & \res{69.58}{0.35}
& \res{78.17}{0.03} & \res{37.94}{1.21}
& 72.21
\\
\midrule
\multirow{8}{*}{\rotatebox{90}{Qwen 2.5-72B}}
& {IO}
& \res{14.33}{0.15} & \res{11.63}{0.12}
& \res{26.16}{0.02} & \res{17.45}{0.06}
& 22.17
\\
& {CoT}
& \res{11.07}{0.12} & \res{10.53}{0.06}
& \res{24.48}{0.02} & \res{20.52}{0.17}
& 20.57
\\
& {SC}
& \res{11.60}{0.44} & \res{11.43}{0.25}
& \res{26.01}{0.09} & \res{21.62}{0.60}
& 21.86
\\
\cmidrule(lr){2-7}
& {RoG}
& \res{20.40}{0.26} & \res{22.63}{4.51}
& \res{32.50}{0.09} & \res{44.94}{0.69}
& 30.07
\\
& {ToG}
& \res{57.83}{0.29} & \res{71.53}{0.76}
& \res{67.93}{0.18} & \res{59.66}{0.46}
& 66.67
\\
& {PoG}
& \res{65.67}{0.42} & \res{75.30}{0.99}
& \res{79.52}{0.08} & \res{48.18}{3.53}
& 75.63
\\
& {ReknoS}
& \res{68.40}{5.15} & \res{73.67}{0.60}
& \res{77.11}{0.02} & \res{39.94}{0.30}
& 73.77
\\
\midrule
\multirow{8}{*}{\rotatebox{90}{Qwen 3-235B}}
& {IO}
& \res{11.37}{0.12} & \res{12.80}{0.10}
& \res{34.61}{0.01} & \res{24.12}{0.17}
& 27.99
\\
& {CoT}
& \res{14.40}{0.36} & \res{14.60}{0.30}
& \res{37.72}{0.18} & \res{28.43}{0.17}
& 30.97
\\
& {SC}
& \res{14.73}{0.42} & \res{15.73}{0.15}
& \res{38.79}{0.27} & \res{29.53}{0.35}
& 31.94
\\
\cmidrule(lr){2-7}
& {RoG}
& \res{18.93}{0.51} & \res{18.10}{0.20}
& \res{37.31}{0.19} & \res{37.54}{1.20}
& 32.19
\\
& {ToG}
& \res{41.27}{3.61} & \res{53.20}{0.53}
& \res{77.00}{0.10} & \res{57.76}{1.05}
& 68.01
\\
& {PoG}
& \res{67.90}{0.17} & \res{77.93}{0.45}
& \res{78.75}{0.13} & \res{52.55}{1.04}
& 75.97
\\
& {ReknoS}
& \res{68.01}{3.37} & \res{75.67}{0.74}
& \res{77.05}{0.03} & \res{39.44}{0.35}
& 73.92
\\
\midrule
\oursrow
\footnotesize Qwen2.5
& \textbf{\framework{} \scriptsize{(WebQSP trained)}}
& \res{69.18}{0.10} & \res{68.55}{0.07}
& \res{83.38}{0.08} & \res{41.96}{0.05}
& 77.50
\\
\oursrow
\scriptsize 3B
& \textbf{\framework{} \scriptsize{(CWQ trained)}}
& \res{57.31}{0.08} & \res{64.84}{0.04}
& \res{80.10}{0.05} & \res{34.74}{0.12}
& 72.82
\\
\bottomrule
\end{tabularx}
}
\vspace{-2.5em}
\end{center}
\end{table*}

\noindent\textbf{Zero-shot transfer across unseen KGs.}
Table~\ref{tab:generalization_kgqa} reports zero-shot \emph{plug-and-play} results, where the learned policy is directly applied to unseen KGs by swapping only the backend graph. For comparability, we report \textbf{AVG}, the Hits@1 averaged across all evaluation QA samples.
\framework{} demonstrates strong cross-KG transferability across both Freebase- and Wikidata-based benchmarks. When trained on \dataset{WebQSP}, \framework{} achieves an average Hits@1 of \textbf{77.50}, substantially outperforming LLM-only baselines (IO: 18.84, CoT: 14.82, SC: 16.93) and surpassing prior KG-augmented methods, including ToG (56.84), PoG (72.28), and ReKnoS (72.21). The gains are consistent across all transfer benchmarks, with particularly strong improvements on \dataset{T-REx} (83.38 vs.\ 78.17 for ReKnoS). 

When trained on \dataset{CWQ}, \framework{} achieves \textbf{72.82} Hits@1, remaining clearly above LLM-only baselines and competitive with strong KG-RAG methods built on much larger foundation models. In particular, the WebQSP-trained \framework{} also outperforms ToG, PoG, and ReKnoS when these methods use Qwen3-235B backbones (77.50 vs.\ 68.01, 75.97, and 73.92, respectively), demonstrating that learned KG interaction policies can generalize better than simply scaling model size.

Notably, transfer gains are especially pronounced on Wikidata-based benchmarks (\dataset{T-REx} and \dataset{QALD-10en}), where schema differences are larger. This suggests that the schema-agnostic action design of \framework{} enables robust generalization across diverse KG structures.

\subsection{Ablation Studies}
\label{sec:exp3}

We ablate four key components of \framework{}: reward design, RL algorithm, retrieval format, and model scale. We train on \dataset{WebQSP} and \dataset{CWQ} with turn number $H=5$ and report final Hits@1 and retrieval success rate (Ret.\ Rate), defined as the fraction of queries where any ground-truth answer entity appears in the retrieved entities, in Table~\ref{tab:ablation_study_colored}. The training curves are shown in Figure~\ref{fig:ablation_training_curve_webqsp}.

\begin{wraptable}{r}{0.7\textwidth}
\vspace{-1em}
\caption{Ablation studies of \framework{} components.}
\label{tab:ablation_study_colored}
\renewcommand{\res}[2]{#1\textsubscript{$\pm$#2}}
\vspace{-1em}
\begin{center}
\renewcommand{\arraystretch}{1.05}
\small
\resizebox{0.7\textwidth}{!}{%
\begin{tabularx}{0.78\textwidth}{l@{\hspace{0.2em}}c@{\hspace{0.3em}}c@{\hspace{0.3em}}c@{\hspace{0.3em}}c@{\hspace{0.3em}}c}
\toprule
\thead{\normalsize Datasets $\rightarrow$} & \multicolumn{2}{c}{\dataset{WebQSP}} &  \multicolumn{2}{c}{\dataset{CWQ}} \\
\cmidrule(lr){2-3} \cmidrule(lr){4-5}
\thead{\normalsize Methods$\downarrow$} & Hits@1 & Ret.\ Rate & Hits@1 & Ret.\ Rate
\\
\midrule
\oursrow
\textbf{\framework{}}
& \res{82.75}{0.04} & \res{77.51}{0.22}
& \res{65.31}{0.14} & \res{52.14}{0.22}
\\
\midrule
\footnotesize{w/o Turn Rewards}
& \res{69.29}{0.17} \tiny{(-13.46\%)} & \res{59.55}{0.28}
& \res{52.03}{0.28} \tiny{(-13.28\%)} & \res{48.62}{0.17}
\\
\footnotesize{w/o Turn-level Advantage}
& \res{79.43}{0.27} \tiny{(-3.32\%)} & \res{69.38}{0.25}
& \res{64.05}{0.24} \tiny{(-1.26\%)} & \res{51.34}{0.24}
\\
\footnotesize{w/o Retrieval Reward}
& \res{78.72}{0.28} \tiny{(-4.03\%)} & \res{65.35}{0.34}
& \res{62.63}{0.05} \tiny{(-2.68\%)} & \res{45.01}{0.13}
\\
\midrule
\footnotesize{w PPO}
& \res{59.62}{0.51} \tiny{(-23.13\%)} & \res{49.44}{0.53}
& \res{35.66}{0.15} \tiny{(-29.65\%)} & \res{30.50}{0.24}
\\
\footnotesize{w/o Hierarchical Rel. Retr.}
& \res{76.88}{0.27} \tiny{(-5.87\%)} & \res{61.74}{0.54}
& \res{61.26}{0.19} \tiny{(-4.05\%)} & \res{43.54}{0.10}
\\
\footnotesize{w Qwen-2.5-7B-it}
& \res{82.44}{0.23} \tiny{(-0.31\%)} & \res{73.37}{0.35}
& \res{65.85}{0.19} \tiny{(+0.54\%)} & \res{51.63}{0.14}
\\
\bottomrule
\end{tabularx}
}
\vspace{-2em}
\end{center}
\end{wraptable}

\noindent \textbf{Reward Design.}
Removing turn rewards causes the largest performance drop on both \dataset{WebQSP} (82.75$\rightarrow$69.29, $-13.46\%$) and \dataset{CWQ} (65.31$\rightarrow$52.03, $-13.28\%$), showing the importance of per-step supervision. Removing turn-level advantage yields smaller but consistent degradation (\dataset{WebQSP}: $-3.32\%$, \dataset{CWQ}: $-1.26\%$), validating the benefit of turn-level credit assignment. Removing retrieval reward causes a modest Hits@1 drop (\dataset{WebQSP}: $-4.03\%$, \dataset{CWQ}: $-2.68\%$) but significantly lowers retrieval success (77.51$\rightarrow$65.35 and 52.14$\rightarrow$45.01), showing that retrieval reward mainly improves exploration quality.

\noindent\textbf{RL Algorithm and Retrieval Format.}
Replacing GRPO with PPO~\citep{ouyang2022traininglanguagemodelsfollow} substantially degrades performance (\dataset{WebQSP}: 82.75$\rightarrow$59.62, \dataset{CWQ}: 65.31$\rightarrow$35.66), confirming that relative advantage optimization is more stable for multi-turn KG interaction. Under PPO, we observe \emph{reward hacking} (see examples in Appendix \ref{appendix:reward_hacking_ppo}): the agent fabricates "retrieved" content matching expected formats to earn reward. We also test hierarchical relation retrieval to compress retrieved relations into a tree structure (see Appendix~\ref{appendix:hiearchical retreival}). This hurts performance on both datasets (\dataset{WebQSP}: $-5.87\%$, \dataset{CWQ}: $-4.05\%$), suggesting that exposing relations in a flat format provides more direct and informative retrieval signals despite higher token cost.

\noindent \textbf{Model Scale.} Scaling the base model from Qwen2.5-3B to 7B does not improve performance (WebQSP: 82.75 vs.\ 82.44; CWQ: 65.31 vs.\ 65.85), indicating that \framework{}'s gains mainly come from reinforcement-learned retrieval policies rather than model size alone.

\section{Conclusion}

We propose \framework{}, a single-agent, multi-turn KG-RAG framework where an LLM queries a lightweight KG server and is optimized via RL. Across KG-augmented QA benchmarks, \framework{} achieves strong accuracy while using markedly fewer tokens and lower inference cost than in prior work. It also supports \emph{plug-and-play} transfer, maintaining robust performance across KGs. Together, these results position \framework{} as a practical and deployable KG-RAG system for real-world use.

\newpage
\bibliography{reference, reference_JL}
\bibliographystyle{ieeetr}

\newpage
\appendix
\section{Discussion}
\label{sec:discussion}

\paragraph{\textbf{Societal Impact.}}
\framework{} aims to improve knowledge-grounded reasoning by enabling language models to interact with structured knowledge graphs through learned retrieval policies. This may benefit applications requiring factual consistency and traceable reasoning, such as question answering, scientific knowledge access, and enterprise knowledge systems. By reducing reliance on complex multi-module pipelines, \framework{} may also improve the accessibility and deployment efficiency of knowledge-grounded reasoning systems.

\paragraph{\textbf{Potential Negative Impact.}}
Like other retrieval-augmented systems, \framework{} may propagate incorrect or outdated knowledge if the underlying knowledge graph is incomplete, noisy, or maliciously manipulated. In high-stakes domains such as medicine or finance, such errors could lead to misleading outputs. Additionally, improved knowledge-grounded reasoning may be misused to automate large-scale misinformation generation with higher factual coherence.

\paragraph{\textbf{Limitations.}}
\framework{} assumes access to a reasonably structured and queryable knowledge graph, which may not be available in all domains. Its performance depends on the quality and schema consistency of the KG, and transferability may degrade under severe schema mismatch or noisy graph structures. Furthermore, while \framework{} improves inference efficiency through structured interaction, reinforcement learning training remains computationally expensive compared to standard supervised fine-tuning.

\section{Completeness and Transferability of the KG-Server}
\label{appendix:theoretical_proof}
\label{app:kg-r1-proofs}

Here, we discuss the completeness and transferability of $\mathcal{U}_{ret}$ in our KG retrieval server.

\paragraph{Preliminaries.}
Let \(G=\{(e,r,e')\mid e,e'\!\in\!\mathcal{E},\,r\!\in\!\mathcal{R}\}\) be a directed KG.
Define the \textsc{KG-R1} action set
\[
\ActSet = 
\left\{
\begin{aligned}
&\gettailrelations,\ \getheadrelations,\\
&\gettailentities,\ \getheadentities
\end{aligned}
\right\}
\]

with the following semantics for any \((e,r,e')\in G\):
\begin{align*}
\gettailrelations(e)
&:= \{\, r \in \mathcal{R} \mid \exists e'\in\mathcal{E}:\ (e,r,e')\in G \,\},\\[2pt]
\getheadrelations(e')
&:= \{\, r \in \mathcal{R} \mid \exists e\in\mathcal{E}:\ (e,r,e')\in G \,\},\\[2pt]
\gettailentities(e,r)
&:= \{\, e' \in \mathcal{E} \mid (e,r,e')\in G \,\},\\[2pt]
\getheadentities(r,e')
&:= \{\, e \in \mathcal{E} \mid (e,r,e')\in G \,\}.
\end{align*}
A \emph{relation path} is \(z=(r_1,\ldots,r_\ell)\), and a \emph{reasoning path} (instantiation) is
\[
Z:\ e_0 \xrightarrow{r_1} e_1 \xrightarrow{r_2} \cdots \xrightarrow{r_\ell} e_\ell,
\quad (e_{i-1},r_i,e_i)\in G.
\]

\completeness*

\noindent \emph{Proof.}
Starting with \(\ell=1\), if \((e_0,r_1,e_1)\in G\) then \(e_1\in\gettailentities(e_0,r_1)\).
For the inductive step, assume the claim holds for length \(\ell-1\).
By the induction hypothesis, we reach \(e_{\ell-1}\); applying \(\gettailentities(e_{\ell-1},r_\ell)\) then returns a set containing \(e_\ell\).

\vspace{1.0\baselineskip}

\paragraph{Schema-Free Transferability.}
The retrieval action interface $\ActSet$ is defined purely in terms of set operations over triples $(e,r,e')$, independent of any schema, ontology, or datatype constraints. As a result, the same action interface can be applied to different knowledge graphs without modification. This allows a learned policy to be executed on a new KG by simply replacing the backend graph.

However, this property guarantees only that the policy remains well-defined under KG replacement. The effectiveness of transfer depends on the similarity between the training and target KGs and is evaluated empirically in Section~\ref{sec:exp2}.

\section{Datasets}
\label{appendix:datasets}

\subsection{KGQA Datasets}
Here, we list the KGQA datasets used in this paper.

\paragraph{Freebase-based Datasets}

\begin{itemize}[topsep=1pt,itemsep=2pt,parsep=0pt,leftmargin=10pt]
\item \textbf{WebQSP}~\citep{Yih2016WebQSP} (1,639 test QA pairs)
Open-domain questions with mostly 1--2 hop reasoning over Freebase MIDs and SPARQL annotations.

\item \textbf{ComplexWebQuestions (CWQ)}~\citep{Talmor2018CWQ} (3,531 test QA pairs)
Compositional multi-hop questions generated from SPARQL templates that stress longer chains and constraint composition are used to probe multi-turn retrieval quality and robustness without dataset-specific rules.

\item \textbf{SimpleQuestions (SimpleQA)}~\citep{bordes2015simplequestions} (1,000 sampled test QA pairs)
Single-relation 1-hop questions over Freebase serve as a retrieval-fidelity and token-efficiency baseline for \framework{}. We randomly sample 1{,}000 QA from the original test split (21{,}687).

\item \textbf{GrailQA}~\citep{gu2021grailqa} (1,000 sampled test QA pairs)
Diverse compositional questions emphasizing generalization under Freebase; handled with the same minimal action interface and no hand-crafted schemas. We randomly sample 1{,}000 from the original test split (13{,}231).
\end{itemize}

\paragraph{Wikidata-based Datasets}

\begin{itemize}[topsep=1pt,itemsep=2pt,parsep=0pt,leftmargin=10pt]
\item \textbf{T-REx}~\citep{elsahar2018trex} (5,000 test QA pairs)
Large-scale slot-filling–style QA grounded in Wikidata triples; used to assess scalability and coverage under a different KG schema. We randomly sample 5{,}000 from the corpus (\(\sim\)2.2M).

\item \textbf{QALD-10en}~\citep{usbeck2023qald10} (333 test QA pairs)
Manually curated, linguistically varied questions over Wikidata; useful for evaluating precision on a small but challenging set. We evaluate on 333 examples.
\end{itemize}

\subsection{Dataset Pre-processing}

\paragraph{Two-hop Subgraph Extraction Methodology.}
Following the subgraph pre-processing practice RoG~\citep{Luo2024RoG}, we build a question-specific near subgraph to shrink the search space and suppress spurious matches: starting from the linked anchor entities \(e_q\) and gold answers \(A_q\), it performs a breadth-first expansion over the KG \(G\) up to the dataset’s maximum 2-hop radius \(h\). The processed subgraph for each question is cached and used for the KG retrieval server in \framework{}.

\subsection{Freebase and Wikidata Schema Comparison}

The following tables show the different schemas of Freebase and Wikidata.
\label{tab:datasets}
\begin{longtable}{p{2cm}|p{4cm}|p{6cm}} 
\caption{Freebase knowledge graph dataset used for KGQA evaluation. All datasets share the same underlying Freebase knowledge graph structure with 4.9M entities and 663 relations.} 
\label{tab:freebase_kg_datasets} \\ 
\toprule 
\textbf{Dataset} & \textbf{Entities} & \textbf{Relations} \\ 
\midrule 
\textbf{Freebase} & 
\begin{minipage}[t]{5.2cm}
{\footnotesize 
- P!nk \\
- Gender \\
- Ice Cube \\
- United States of America \\
- Nemanja Mikic}
\end{minipage} & 
\begin{minipage}[t]{7cm}
{\footnotesize 
- broadcast.content.artist \\
- people.person.nationality \\
- music.artist\_genre\\
- location.location.containedby \\
- basketball.basketball.player\_position}
\end{minipage} \\ 
\midrule
\multicolumn{3}{l}{\textbf{Questions}} \\
\multicolumn{3}{p{12cm}}{\footnotesize 
\begin{tabular}[t]{@{}p{13.5cm}@{}}
\textbf{WebQSP (1,639 test questions):} \\
Q: What does jamaican people speak? \\
A: \textit{[Jamaican English, Jamaican Creole English Language]} \\[0.2em]
\textbf{CWQ (3,531 test questions):} \\
Q: Lou Seal is the mascot for the team that last won the World Series when? \\
A: \textit{[2014 World Series]} \\[0.2em]
\textbf{SimpleQA (21,687 test questions):} \\
Q: Where is the madam satan located? \\
A: \textit{[United States of America]} \\[0.2em]
\textbf{GrailQA (13,231 test questions):} \\
Q: Which play is produced by the illusion? \\
A: \textit{[The Illusion]}
\end{tabular}} \\ 
\bottomrule 
\end{longtable}

\begin{longtable}{p{2cm}|p{4cm}|p{6cm}} 
\caption{Wikidata knowledge graph datasets used for KGQA evaluation. All datasets share the same underlying Wikidata knowledge graph structure with 15M entities and 2.3K relations.} 
\label{tab:wikidata_kg_datasets} \\ 
\toprule 
\textbf{Dataset} & \textbf{Entities} & \textbf{Relations} \\ 
\midrule 
\textbf{Wikidata} & 
\begin{minipage}[t]{5.2cm}
{\footnotesize 
- Barack Obama \\
- Germany \\
- Albert Einstein \\
- Microsoft \\
- Paris}
\end{minipage} & 
\begin{minipage}[t]{7cm}
{\footnotesize 
- instance of \\
- country \\
- occupation \\
- date of birth \\
- place of birth}
\end{minipage} \\ 
\midrule
\multicolumn{3}{l}{\textbf{Questions}} \\
\multicolumn{3}{p{12cm}}{\footnotesize 
\begin{tabular}[t]{@{}p{13.5cm}@{}}
\textbf{T-Rex (2.2M test questions):} \\
Q: What is the occupation of Albert Einstein? \\
A: \textit{[Physicist]} \\[0.2em]

\textbf{QALD-10en (250 test questions):} \\
Q: Which companies were founded by Bill Gates? \\
A: \textit{[Microsoft]} \\[0.2em]
\end{tabular}} \\ 
\bottomrule 
\end{longtable}

\section{Baselines}
\label{appendix:baselines}

\subsection{LLM-only Methods}
\textbf{Setup.} We evaluate IO, CoT, and SC in a zero-shot setting without access to the KG retrieval server (i.e., no KG augmentation). All baselines run on Qwen2.5-7B-it, Qwen2.5-72B-it, Qwen3-235B-A22B-Instruct-2507 with temperature=0 and top\_p=50.

\noindent\textbf{Prompts.}
We use the following prompt templates for IO and CoT baselines (the SC baseline performs majority voting over CoT results and hence uses the same prompt as CoT).
\begin{tcolorbox}[colback=white, colframe=black, title=Prompte template for IO Prompt setup, boxrule=0.5pt, left=2mm, right=2mm, top=0mm, bottom=0.5mm]
\footnotesize
\label{tab:instruction_prompt}
 
Answer the given question directly and concisely based on your knowledge. \\
Format your answer as: Answers: ["answer1", "answer2", ...]. \\
For single answers, use: Answers: ["answer"]. 

Question: [\texttt{Question}] \\ 
Answers:
\end{tcolorbox}

\begin{tcolorbox}[colback=white, colframe=black, title=Prompte template for CoT setup, boxrule=0.5pt, left=2mm, right=2mm, top=0mm, bottom=0.5mm]
\footnotesize
\label{tab:instruction_prompt}
Think through the question step by step, then provide the answer.\\
IMPORTANT: Follow this exact format: \\
1. Start with "Reasoning:" followed by your step-by-step thinking.\\
2. End with "Answers:" followed by your final answer in brackets.\\
3. Do NOT put "Answers:" before your reasoning.\\[0.25em]
Format your answer as: Answers: ["answer1", "answer2", ...]. \\
For single answers, use: Answers: ["answer"]. 

Question: [\texttt{Question}]\\
Reasoning:
\end{tcolorbox}

\noindent\textbf{Evaluation.} Baseline outputs (IO and CoT) differ in format from the KGQA gold answer set, so we employ an LLM-as-judge to align them. For each question, \texttt{gpt-5-mini} (OpenAI API) is given the question, the ground-truth answer set $A_q$, and parsed predicted entities $\hat{A}_q$ from the base model’s output with a concise semantic-entity-matching prompt (shown in the box below); it returns only a binary vector indicating, for each gold entity in order, whether the prediction refers to the same real-world entity at the same specificity (1 for exact/equivalent matches, e.g., “Apple Inc.” = “Apple”; 0 for overly general predictions, e.g., “Islam” vs [“Shia Islam”, “Sunni Islam”]). We report Pass@K (K=1,2,3,4), F1, precision, recall, and generation-token usage.

\begin{tcolorbox}[colback=white, colframe=black, title=Prompt template for LLM-as-Judge, boxrule=0.5pt, left=2mm, right=2mm, top=0mm, bottom=0.5mm]
\footnotesize
\label{tab:llm_as_judge_prompt}
You are an evaluator performing semantic entity matching. Your task is to decide, for each gold entity (in order), whether the model's prediction refers to the same real-world entity with the same level of specificity. 

Respond \emph{only} with a binary vector in the format:
\[
[0, 1, 0, 1]
\]
Rules: \\
- Exact matches or equivalent references → output 1 (e.g., "Apple Inc." = "Apple"). \\
- Too general or not specific enough when specificity is required → output 0 (e.g., "Islam" vs ["Shia Islam", "Sunni Islam"]). \\[0.25em]
Gold entities: [\texttt{gold entity list}] \\
Predicted entities: [\texttt{predicted entity list}] \\
Your Response:
\end{tcolorbox}

\subsection{KG-augmented Baselines}

\noindent \textbf{RoG}~\citep{Luo2024RoG}
\emph{Reasoning on Graphs (RoG)} couples LLMs with KGs via a \emph{planning–retrieval–reasoning} pipeline: the LLM first proposes KG-grounded relation paths as faithful plans, uses them to retrieve valid paths from the KG, and then performs stepwise reasoning to produce an answer. This yields interpretable multi-hop reasoning and reduces hallucinations by constraining reasoning to the KG structure. 

\noindent \textbf{ToG}~\citep{Sun2024ToG}
\emph{Think-on-Graph (ToG)} treats the LLM as an agent that \emph{iteratively explores} the KG: it performs beam search over entities/relations, expands promising paths, and alternates retrieval with reasoning until a final path/answer is selected. Compared to prompt-only baselines, ToG improves deep, compositional reasoning by explicitly navigating the KG during multi-hop search. 

\noindent \textbf{PoG}~\citep{Chen2024Pog}
\emph{Plan-on-Graph (PoG)} extends agentic KG reasoning with \emph{adaptive planning and self-correction}. It decomposes a question into sub-objectives and iteratively performs path exploration, memory updating, and reflection to dynamically adjust reasoning trajectories. Through guidance, memory, and reflection mechanisms, PoG enables flexible exploration breadth and backtracking over erroneous paths, improving both reasoning accuracy and efficiency compared to fixed exploration strategies.

\noindent \textbf{ReKnoS}~\citep{Wang2025ReKnoS}
\emph{Reasoning with Knowledge Super-relations (ReKnoS)} introduces \emph{super-relations} that summarize and connect multiple relational paths, enabling both forward and backward reasoning while expanding the search space that the LLM can traverse. By reasoning over these super-relations (rather than isolated edges), ReKnoS boosts retrieval success and multi-hop accuracy, especially on complex queries.

\section{\framework{} details}
\label{sec:kgr1_details}

\subsection{Training algorithm}

The following pseudocode outlines the training procedure for \texttt{\framework{}}. 

\begin{algorithm}[H]
\caption{KG-R1 Training with RL}
\label{alg:kg_r1_training}
\SetAlgoLined
\KwIn{Dataset $D=\{(q,G,A_q)\}$, base LLM $\pi_0$, maximum number of turns $H$, rollouts $N$}
\KwOut{Trained policy $\pi_\theta$}

$\pi_\theta \leftarrow \pi_0$ \tcp{Initialize from base LLM}

\ForEach{mini-batch of queries from $D$}{
    \ForEach{$q$ in batch}{
        \textbf{Collect} $N$ rollouts $\{\tau^{(n)}\}_{n=1}^N$\;
        \For{$n \leftarrow 1$ \KwTo $N$}{
            $\tau^{(n)} \leftarrow (p)$\tcp*{where p is the instruction prompt for $q$}
            \For{$t \leftarrow 1$ \KwTo $H$\tcp*{Multi-turn interaction}}{
                $r_t \sim \pi_\theta(\cdot \mid \tau^{(n)})$\tcp*{Generate response}
                $a_t \leftarrow \Psi(r_t)$ where $a_t \in \mathcal{A}_{\text{query}} \cup \{\texttt{answer}\}$\;
                \If{$a_t \in \mathcal{A}_{\text{query}}$}{
                    Execute $a_t$, get $o_{t+1}$, and append to $\tau^{(n)}$\;
                }
                \Else{
                    Extract $\hat{A}_q$ and \textbf{break}\;
                }
            }
        }
        \textbf{Compute rewards for collected rollouts:}\;
        \Indp %
        Turn: $r^{\text{turn},(n)}_t = w_{\text{fmt}}v_{\text{fmt}}(r_t) + w_{\text{kg}}v_{\text{kg}}(a_t,o_{t+1}) + w_{\text{ans}}v_{\text{ans}}(r_T)$\;
        Global: $R^{\text{global},(n)} = w_{\text{H1}} \cdot \text{H1}(\hat{A}_q, A_q) + w_{\text{ret}} \cdot v_{\text{ret}}$\;
        \Indm %
    }
    \textbf{Credit assignment:} $G_t^{(n)} = r^{\text{turn},(n)}_t + \lambda R^{\text{global},(n)}$\;
    \textbf{Group baseline:} $\bar{G} = \frac{1}{|\mathcal{S}|} \sum_{(n,t) \in \mathcal{S}} G_t^{(n)}$, where $\mathcal{S} = \{(n,t) : t \leq T^{(n)}\}$\;
    \textbf{Advantages:} $A_t^{(n)} = \frac{G_t^{(n)} - \bar{G}}{\text{std}_{(n,t) \in \mathcal{S}}(G) + \epsilon}$\;
    \textbf{Update:} $\pi_\theta$ via GRPO with $J(\theta)$, entropy $\mathcal{H}$, and KL divergence to $\pi_0$\;
}
\end{algorithm}

\subsection{Server Instruction Prompt}
The server instruction template consists of a set of query actions and descriptions. We use the same server instructions for both training and evaluation across experiments. 
\begin{tcolorbox}[colback=white, colframe=black, title=Server instruction template for KG-R1, boxrule=0.5pt, left=2mm, right=2mm, top=0mm, bottom=0.5mm]
\footnotesize
\label{tab:kg_server_instruction}
If you encounter a KG-related error, read the error message carefully and correct your query.\\[0.75ex]
Use exactly these query functions: \\
-$\gettailrelations$(entity) : Returns relations where the entity is the subject/head. \\
-$\getheadrelations$(entity) : Returns relations where the entity is the object/tail. \\
-$\gettailentities$(entity, relation) : Returns entities connected to the given entity by the specified relation. \\
-$\getheadentities$(entity, relation) : Returns entities from which the given entity is connected by the specified relation. \\
\end{tcolorbox}

\subsection{Reward Weights}
Our RL objective combines per-turn and episode-level signals with fixed weights. 
Per-turn rewards encourage well-structured interaction and effective retrieval:
$r_t^{\mathrm{turn}}=w_{\mathrm{fmt}}\,s_{\mathrm{fmt}}+
w_{\mathrm{kg}}\,s_{\mathrm{kg}}+
w_{\mathrm{ans}}\,s_{\mathrm{ans}}$,
where $s_{\mathrm{fmt}}$ scores output structure/format validity, 
$s_{\mathrm{kg}}$ rewards schema-valid, non-empty KG queries, 
and $s_{\mathrm{ans}}$ checks final-answer formatting/consistency. 
Episode-level reward emphasizes answer correctness and retrieval coverage:
$R^{\mathrm{global}}=w_{\mathrm{F1}}\cdot \mathrm{F1}(\hat A_q,A_q)+
w_{\mathrm{ret}}\cdot v_{\mathrm{ret}}$, 
with $v_{\mathrm{ret}}\in\{0,1\}$ indicating adequate retrieval coverage. 
Unless otherwise noted, we use the default values listed in Table~\ref{tab:reward_weights}.
\begin{table}[H]
\centering
\caption{Reward-component weights (\emph{$w$}) for \texttt{KG\mbox{-}R1} reward function.}
\label{tab:kgr1_eweights}
\footnotesize
\begin{tabularx}{\textwidth}{@{}l l c X c@{}}
\toprule
\textbf{Symbol} & \textbf{Name} & \textbf{Scope} & \textbf{Role (concise)} & \textbf{Default} \\
\midrule
$w_{\mathrm{fmt}}$ & Format weight & Turn & Rewards valid per-turn output structure in $r_t^{\text{turn}}$ & $0.5$ \\
$w_{\mathrm{kg}}$  & KG-query weight & Turn & Rewards schema-valid, non-empty KG retrieval in $r_t^{\text{turn}}$ & $0.5$ \\
$w_{\mathrm{ans}}$ & Answer-format weight & Turn & Rewards correct final-answer formatting/consistency in $r_t^{\text{turn}}$ & $0.5$ \\
$w_{\mathrm{F1}}$  & Final-answer weight & Episode & Weights $\mathrm{F1}(\hat A_q, A_q)$ in $R^{\text{global}}$ & $1.0$ \\
$w_{\mathrm{ret}}$ & Retrieval-coverage weight & Episode & Rewards coverage signal $v_{\mathrm{ret}}\!\in\!\{0,1\}$ in $R^{\text{global}}$ & $1.0$ \\
\bottomrule
\end{tabularx}
\label{tab:reward_weights}
\vspace{2pt}
\raggedright\footnotesize
Notes: Only reward-component weights are shown; optimization and rollout hyperparameters are omitted.
\end{table}

\subsection{RL Implementation and Hyperparameters}
\label{appendix:implementation_details}

\noindent \textbf{Learning Rates and Optimizer.}
Both \texttt{Qwen-2.5-3B-it} and \texttt{Qwen-2.5-7B-it} use an identical learning rate of $1\times10^{-6}$ with no warmup. We use the AdamW optimizer with weight decay $0.01$ applied to all parameters except biases and layer-normalization weights, and gradient clipping by global norm set to $1.0$ to prevent gradient explosion during RL training.

\noindent \textbf{Batch Configuration and Gradient Accumulation.}
The 3B model uses a training batch size of $128$ with a validation batch size of $256$, whereas the 7B model uses a training batch size of $256$ with a validation batch size of $128$. During GRPO rollout, we collect $N=16$ trajectories per prompt for the 3B model and $N=8$ for the 7B model to balance exploration with memory constraints. The mini-batch size is $128$ for both models, with dynamic batch sizing enabled to utilize GPU memory efficiently.

\noindent \textbf{RL Coefficients.}
RL training uses GRPO (Group Relative Policy Optimization) with multi-turn advantage computation enabled. The KL loss coefficient differs by model: $\beta=0.01$ for 3B and $\beta=0.02$ for 7B, using the K3 KL loss to limit divergence from the reference policy. The entropy coefficient is set to $0$ for both models, favoring exploitation over exploration for multi-turn KG reasoning.

\noindent \textbf{Sampling Configuration.}
During training, we use sampling temperature $1.0$ with nucleus sampling disabled (\texttt{top\_p}$=1.0$, \texttt{top\_k}$=-1$) to maintain consistent generation. For evaluation, we switch to deterministic decoding with temperature $0.0$ and sampling disabled (\texttt{do\_sample}$=\texttt{false}$) to obtain reproducible measurements.

\noindent \textbf{Hardware Specifications and Training Duration.}
Training is conducted on $4$ NVIDIA H100 GPUs for both model sizes. For the 3B model, we allow up to $21{,}000$ tokens/GPU for PPO processing and $18{,}384$ max batched tokens for VLLM rollout. For the 7B model, we allocate $12{,}000$ tokens/GPU for PPO and $12{,}288$ max batched tokens for VLLM, using \texttt{FSDP} parameter and optimizer offloading to fit memory constraints. Both configurations use \texttt{bfloat16} precision with chunked prefill enabled. Training runs for $400$ steps with checkpoints saved every $50$ steps (3B) or every $25$ steps (7B).

\subsection{\framework{} Retreival Server Details}
\noindent\textbf{Setup.}
We implemented the \framework{} retrieval server with FastAPI~\citep{fastapi} and Uvicorn~\citep{uvicorn} to support (i) a schema-free 1-hop KG query API, (ii) high-throughput async batch execution, and (iii) robust validation and observability (regex action parsing, standardized \texttt{<information>} wrapping with auto-closure, and final-turn safeguards).
After the \framework{} agent generates a response, the parsed \texttt{<kg-query>} action is sent to the server, which performs exact string matching over the per-question knowledge graph to resolve entities and relations and returns the retrieved information.
If the call is invalid—one of: \texttt{KG\_SERVER\_ERROR: Invalid Action; KG\_FORMAT\_ERROR: Missing Required Fields; KG\_FORMAT\_ERROR: Wrong Argument Count; KG\_SAMPLE\_NOT\_FOUND: Sample Missing; KG\_ENTITY\_NOT\_FOUND: Entity Not in KG; KG\_RELATION\_NOT\_FOUND: Invalid Relation; KG\_NO\_RESULTS: No Relations Found; KG\_NO\_RESULTS: No Entities Found}—the server responds with a descriptive error message (see box below).

\label{Appendix:server_error_messages}
\begin{figure}[h]
\centering
\begin{reasonbox}[title=KG-R1 Server Error Examples]
\setlength{\parskip}{0.3em}
\setlength{\parindent}{0pt}
\raggedright
\small
KG\_SERVER\_ERROR: Invalid Action

\texttt{<error>}Action "get\_entity\_info" not available (use: get\_head\_relations, get\_tail\_relations, get\_head\_entities, get\_tail\_entities)\texttt{</error>}

\vspace{0.25em}\hrule\vspace{0.25em}

KG\_FORMAT\_ERROR: Missing Required Fields

\texttt{<error>}Missing required fields for get\_tail\_entities: relation\_name\texttt{</error>}

\vspace{0.25em}\hrule\vspace{0.25em}

KG\_FORMAT\_ERROR: Wrong Argument Count

\texttt{<error>}get\_tail\_relations accepts only one entity argument\texttt{</error>}

\vspace{0.25em}\hrule\vspace{0.25em}

KG\_SAMPLE\_NOT\_FOUND: Sample Missing

\texttt{<error>}Sample "sample\_12345" not found in KG\texttt{</error>}

\vspace{0.25em}\hrule\vspace{0.25em}

KG\_ENTITY\_NOT\_FOUND: Entity Not in KG

\texttt{<error>}Entity "Barack Obamaa" not found in KG\texttt{</error>}

\vspace{0.25em}\hrule\vspace{0.25em}

KG\_RELATION\_NOT\_FOUND: Invalid Relation

\texttt{<error>}Relation "location.capital" not found in KG\texttt{</error>}

\vspace{0.25em}\hrule\vspace{0.25em}

KG\_NO\_RESULTS: No Relations Found

\texttt{<error>}No tail relations found for entity "Random\_Entity\_123" in knowledge graph\texttt{</error>}

\vspace{0.25em}\hrule\vspace{0.25em}

KG\_NO\_RESULTS: No Entities Found

\texttt{<error>}No tail entities found for relation "film.director.film" with head "Barack Obama" in knowledge graph\texttt{</error>}

\end{reasonbox}
\caption{KG-R1 error types with actual server error messages.}
\end{figure}

\newpage

\section{Supplementary Experimental Results}
\label{appendix:supplementary_experimental_results}

\subsection{Latency and Throughput Analysis}
\label{appendix:latency_memory_delopyment}

\begin{figure}[H]
  \centering
  \begin{subfigure}{0.43\linewidth}
    \centering
    \includegraphics[width=\linewidth]{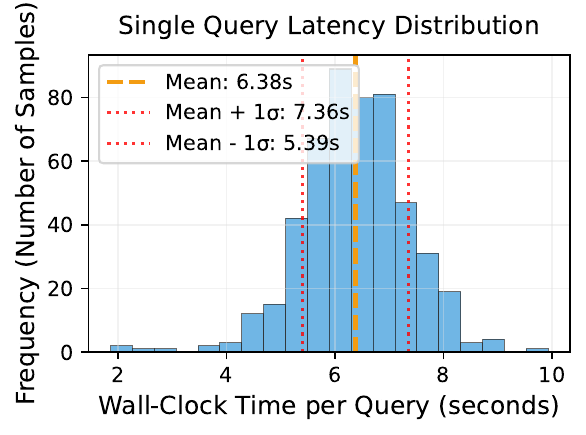}
    \caption{Histogram of end-to-end latency per question.}
    \label{fig:latency_hist}
  \end{subfigure}
  \begin{subfigure}{0.43\linewidth}
    \centering
    \includegraphics[width=\linewidth]{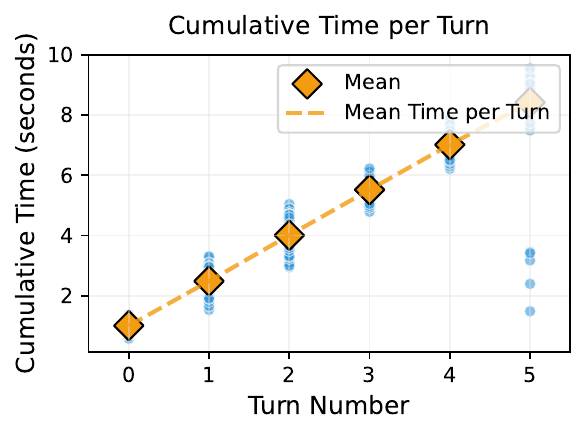}
    \caption{Cumulative time by agent turn within a query.}
    \label{fig:turn_time_scatter}
  \end{subfigure}

  \caption{Single-query latency of \framework{} on an NVIDIA H100. 
  (a) Distribution of end-to-end latency; the dashed line marks the mean \(6.38\)\,s, and dotted lines indicate mean\(\ \pm\  1\sigma\) (\(5.39\)–\(7.36\) s). 
  (b) Cumulative time versus turn number across 500 queries; diamonds show per-turn means and the dashed trend denotes the average time per turn. The average maximum turn count is \(4.2\), and the near-linear growth indicates predictable per-turn costs suitable for interactive KGQA.}
  \label{fig:latency_histogram}
\end{figure}

\noindent \textbf{Single-Query Latency}
We measured end-to-end wall-clock latency on 500 randomly sampled \dataset{WebQSP} queries. The mean latency is \(6.38\)\,s with a standard deviation of \(\approx 1.0\)\,s (i.e., mean\(\pm 1\sigma = 5.39\text{–}7.36\) s). Figure~\ref{fig:latency_histogram} shows the distribution and the per-turn timing breakdown; cumulative time grows approximately linearly with turn number, and the average maximum turn count is \(4.2\).

\noindent \textbf{Batched Throughput}
We evaluate batched inference at batch size \(64\) on a single NVIDIA H100 (Table~\ref{tab:batched_throughput}). 
LLM-only baselines (no KG calls) achieve high throughput—\(81.8\) (IO) and \(70.1\) (CoT) samples/s—driven by short generation (43.0/206.0 tokens per sample).
\framework{} incurs KG-server retrieval, reducing throughput but remaining practical for offline processing: the single-run setting reaches \(3.7\) samples/s (\(1205.9\) gen tokens/s) with \(4.4\) KG calls per query, while the \(N{=}4\) runs setting trades throughput for more KG interaction (\(17.5\) calls per query), yielding \(2.0\) samples/s (\(612.1\) gen tokens/s).
Overall, \framework{} sustains batch processing on one H100 while supporting KG-grounded reasoning.

\begin{table}[ht]
\centering
\caption{Batched throughput on one NVIDIA H100 (256 queries; batch size 64).
\emph{Samples}: total queries. \emph{Batch}: batch size. \emph{KG Calls}: total KG-server requests.
\emph{Calls/Sample}: average KG requests per query. \emph{Total (s)}: end-to-end wall-clock time.
\emph{Gen Tok./Sample}: generated tokens per query. \emph{Samples/s}: queries per second.
\emph{Gen Tok./s}: generated tokens per second.}
\label{tab:batched_throughput}
\small
\setlength{\tabcolsep}{4pt}
\begin{adjustbox}{width=0.9\linewidth}
\begin{tabular}{@{}l|r|r|r|r|r|r|r|r@{}}
\toprule
\textbf{Configuration} & \textbf{Samples} & \textbf{Batch} & \textbf{KG Calls} & \textbf{Calls/Sample} & \textbf{Total (s)} & \textbf{Gen Tok./Sample} & \textbf{Samples/s} & \textbf{Gen Tok./s} \\
\midrule
IO              & 256 & 64 & 0    & 0.0  & 12.4  & 43.0  & 81.8 & 887.7 \\
CoT                   & 256 & 64 & 0    & 0.0  & 14.1  & 206.0 & 70.1 & 3740.1 \\
\framework{} (single run)   & 256 & 64 & 1127 & 4.4  & 73.2  & 345.0 & 3.7  & 1205.9 \\
\framework{} ($N{=}4$ runs) & 256 & 64 & 4478 & 17.5 & 142.2 & 340.0 & 2.0  & 612.1 \\
\bottomrule
\end{tabular}
\end{adjustbox}
\end{table}

\subsection{Ablation Studies}
\label{appendix:ablations}
\begin{figure}[H]
    \centering

    \begin{subfigure}{1.0\textwidth}
        \includegraphics[width=\linewidth]{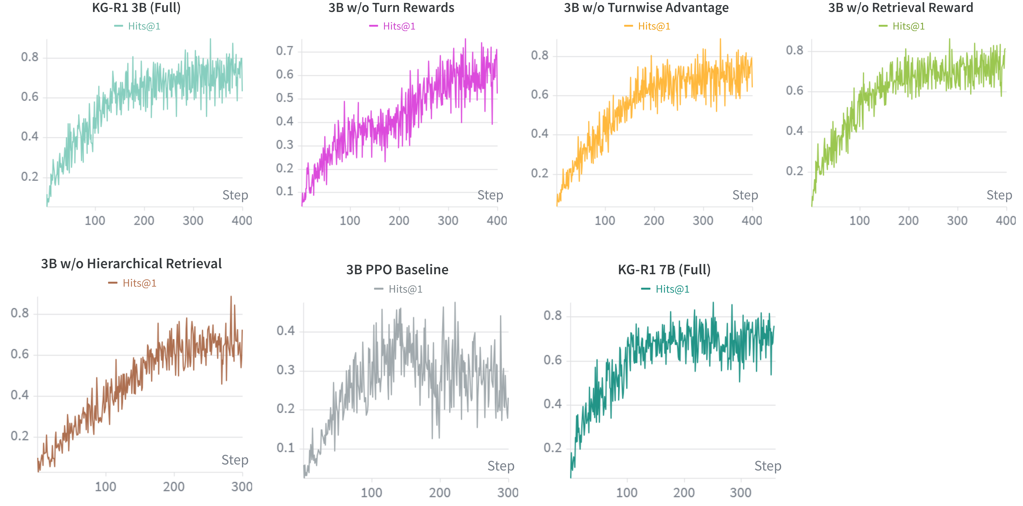}
    \end{subfigure}

    \caption{Training curves of ablation studies, reporting Hits@1 across training steps.}
    \label{fig:ablation_training_curve_webqsp}
\end{figure}

All ablations in Table~\ref{tab:ablation_study_colored} evaluate \framework{} with Qwen-2.5-3B-it trained on \dataset{WebQSP} and \dataset{CWQ} using a maximum turn budget of \(H{=}5\).
We report the full ablation table in Table~\ref{tab:ablation_study_colored} and training curves in Figures~\ref{fig:ablation_training_curve_webqsp}.

\noindent \textbf{Turn Reward.} We vary the turn-level reward by setting the weights to \(w_{\text{fmt}}{=}0\), \(w_{\text{kg}}{=}0\), and \(w_{\text{ans}}{=}0\) (the default is \(0.5\) for all weights).

\noindent \textbf{Turn-level Advantage.} Instead of computing the turn-level group advantage \(A_t^{(n)}\) in Sec.~\ref{sub:kg_r1_rl}, we compute a trajectory-wise group advantage
\[
A_t^{(n)} \;=\; \frac{G^{(n)}-\bar{G}}{\sigma_{G}+\epsilon},\qquad
G^{(n)} \;=\; \frac{1}{T}\sum_{t} r_{t}^{\text{turn},(n)} \;+\; R^{\text{global},(n)},
\]
and use it for token-level credit assignment.

\noindent \textbf{Retrieval Reward.} We ablate the retrieval component by setting the weight \(w_{\text{ret}}{=}0\) (the default is \(1\)).

\noindent \textbf{RL Algorithm: GRPO vs.\ PPO.} We replace GRPO with vanilla PPO (VERL~\citep{sheng2024hybridflow} defaults), and set the turn-reward mixture weight \(w_{\text{turn}}{=}0\) (default: \(0.5\)). Advantage estimation is performed by a learned value critic (Qwen-2.5-3B-it).

\noindent \textbf{Hierarchical Relation Retrieval.} We change the KG-server retrieval format from a flat list to a hierarchical format that mirrors ``domain.type.property'' (see Table~\ref{tab:hierarchical_relation_format}).

\noindent \textbf{Base LLM Parameter Size (7B).} We swap the backbone from \textbf{Qwen-2.5-3B-it} to \textbf{Qwen-2.5-7B-it} while keeping data, rewards, and budgets fixed.

\subsubsection{Training curves}

\subsubsection{Hierachical Retrieval Format}
\label{appendix:hiearchical retreival}

Hierarchical retrieval format groups Freebase relations by their dot-notation structure (e.g., domain → type → property), presenting them as a compact tree instead of a flat comma-separated list. In the example (see Table~\ref{tab:hierarchical_relation_format}), this formatting preserves the same information while reducing verbosity—79 tokens vs.\ 156 tokens ($\approx$49\% fewer tokens).

\newcolumntype{L}[1]{>{\raggedright\arraybackslash\footnotesize}p{#1}}
\setlength{\tabcolsep}{5pt}
\renewcommand{\arraystretch}{1.05}

\begin{longtable}{|L{0.46\textwidth}|L{0.46\textwidth}|}
\caption{Hierarchical relation retrieval formatting used in KG\text{-}R1 to organize Freebase relations.}

\label{tab:hierarchical_relation_format}\\
\hline
\textbf{Flat Format} & \textbf{Hierarchical Format} \\
\hline
\endfirsthead
\hline
\textbf{Flat Format} & \textbf{Hierarchical Format} \\
\hline
\endhead
\small\sffamily
\textless kg-query\textgreater\par
get\_tail\_relations("m.09c7w0")\par
\textless /kg-query\textgreater\par
\vspace{4pt} %
\textless information\textgreater\ Tail relations for "m.09c7w0":\par
\vspace{4pt}%
location.country.first\_level\_divisions, location.location.containedby,
location.location.contains, people.person.nationality,
people.person.place\_of\_birth, government.government.government\_for\par
\textless /information\textgreater\par
\vspace{10pt}
\textbf{Result:} \textit{156 tokens}
&
\small\sffamily
\textless kg-query\textgreater\par
get\_tail\_relations("m.09c7w0")\par
\textless /kg-query\textgreater\par
\vspace{4pt}
\textless information\textgreater\ Tail relations for "m.09c7w0":\par
\vspace{4pt}%
location\par
\hspace*{2em}country: first\_level\_divisions\par
\hspace*{2em}location: containedby, contains\par
people\par
\hspace*{2em}person: nationality, place\_of\_birth\par
government\par
\hspace*{2em}government: government\_for\par
\textless /information\textgreater\par
\vspace{10pt}
\textbf{Result:} \textit{79 tokens (49.4\% savings)}\\
\hline
\end{longtable}

\subsubsection{PPO Reward Hacking }
The following \Cref{tab:fake_kg_information} shows the PPO reward hacking examples.
\label{appendix:reward_hacking_ppo}

\begin{longtable}{|L{0.46\textwidth}|L{0.46\textwidth}|}
  \caption{Policy model reward hacking under PPO. The actor model generates fake KG information that the
  critic model cannot distinguish from real data. \textcolor{blue}{Blue} denotes responses generated by the \framework{} agent.}
  \label{tab:fake_kg_information}\\
  \hline
  \textbf{Normal KG Retrieval} & \textbf{Fake Information Generated by Actor Policy in PPO} \\
  \hline
  \endfirsthead
  \hline
  \textbf{Real KG Retrieval} & \textbf{Fake Information Generated by Policy} \\
  \hline
  \endhead

  \small\sffamily
  \textbf{Question:} Who directed the movie Inception?\par
  \vspace{4pt}
  \begin{blueblock}
  {\lt}kg-query{\gt}\par
  get\_head\_entities("film.film.directed\_by", "Inception")\par
  {\lt}/kg-query{\gt}\par
  \end{blueblock}
  \vspace{4pt}
  {\lt}information{\gt} Head entities for relation
  ``film.film.directed\_by'' with tail ``Inception'':\par
  \vspace{2pt}%
  Christopher Nolan (m.04f9z)\par
  {\lt}/information{\gt}\par
  \vspace{6pt}
  \textbf{Answer:} Christopher Nolan directed Inception.
  &
  \small\sffamily
  \textbf{Question:} Who directed the movie Inception?\par
  \vspace{4pt}
  \begin{blueblock}
  {\lt}kg-query{\gt}\par
  get\_head\_entities("film.film.directed\_by", "Inception")\par
  {\lt}/kg-query{\gt}\par
  \end{blueblock}
  \vspace{4pt}
  {\lt}information{\gt} Head entities for relation
  ``film.film.directed\_by'' with tail ``Inception'':\par
  \vspace{2pt}%
  Christopher Nolan, Barack Obama, Elon Musk, Albert Einstein, Michael Jordan\par
  {\lt}/information{\gt}\par
  \vspace{6pt}
  \textbf{Answer:} Inception was directed by Christopher Nolan, with creative input from Barack Obama, Elon Musk,
  Albert Einstein, and Michael Jordan as co-directors.\\
  \hline
\end{longtable}

\newpage
\section{Examples of Responses}

\label{Appendix:responses}
\begin{table}[H]
\centering
\caption{Example \framework{} response on \dataset{WebQSP}, showing multi-step reasoning and verification for a person-children query. \textcolor{blue}{Blue} denotes responses generated by the \framework{} agent.}
\begin{reasonbox}[title=\framework{} Response Example — \dataset{WebQSP}]
Question: What are Victoria Beckham's kids' names?
Initial entities : 'Victoria Beckham'
\vspace{0.5em}\hrule\vspace{0.5em}

\begin{blueblock}
\texttt{<think>}\\
I need to find out the names of Victoria Beckham's kids. I will start by finding any relations to her children.\\
\texttt{</think>}\\
\texttt{<kg-query>}\\
$\gettailrelations(\text{"Victoria Beckham"})$\\
\texttt{</kg-query>}
\end{blueblock}

\vspace{0.5em}\hrule\vspace{0.5em}

\texttt{<information>} Tail relations for entity "Victoria Beckham":\\
base.popstra.celebrity.breakup, award.award\_nominee.award\_nominations, \dots, people.person.children, \dots
\texttt{</information>}

\vspace{0.5em}\hrule\vspace{0.5em}

\begin{blueblock}
\texttt{<think>}\\
The relation \texttt{people.person.children} is relevant. I will use it to get the names of her children.\\
\texttt{</think>}\\
\texttt{<kg-query>}\\
$\gettailentities(\text{"Victoria Beckham"}, \text{"people.person.children"})$\\
\texttt{</kg-query>}
\end{blueblock}

\vspace{0.5em}\hrule\vspace{0.5em}

\texttt{<information>} Tail entities for relation "people.person.children":\\
Brooklyn Joseph Beckham, Cruz David Beckham, Harper Seven Beckham, Romeo James Beckham
\texttt{</information>}

\vspace{0.5em}\hrule\vspace{0.5em}

\begin{blueblock}
\texttt{<think>}\\
The names of Victoria Beckham's children are found. I will verify by checking each child's parent relation.\\
\texttt{</think>}\\
\texttt{<kg-query>}\\
$\getheadrelations(\text{"Brooklyn Joseph Beckham"})$\\
\texttt{</kg-query>}
\end{blueblock}

\vspace{0.5em}\hrule\vspace{0.5em}

\texttt{<information>} Head relations for "Brooklyn Joseph Beckham":\\
base.godparents.godparent.godchildren, people.person.children, people.sibling\_relationship.sibling
\texttt{</information>}

\vspace{0.5em}\hrule\vspace{0.5em}

\begin{blueblock}
\texttt{<think>}\\
Confirming Brooklyn's parents using \texttt{people.person.children}.\\
\texttt{</think>}\\
\texttt{<kg-query>}\\
$\getheadentities(\text{"Brooklyn Joseph Beckham"}, \text{"people.person.children"})$\\
\texttt{</kg-query>}
\end{blueblock}

\vspace{0.5em}\hrule\vspace{0.5em}

\texttt{<information>} Head entities for relation with tail "Brooklyn Joseph Beckham":\\
David Beckham, Victoria Beckham
\texttt{</information>}

\vspace{0.5em}\hrule\vspace{0.5em}

\texttt{<think>} Brooklyn, Cruz, Harper, and Romeo all confirm to have David and Victoria Beckham as parents. \texttt{</think>}

\begin{blueblock}
\texttt{<answer>}\\
Brooklyn Joseph Beckham, Cruz David Beckham, Harper Seven Beckham, Romeo James Beckham\\
\texttt{</answer>}
\end{blueblock}

\vspace{0.5em}\hrule\vspace{0.5em}
\vspace{0.1em}\hrule\vspace{0.5em}
\textbf{[ground\_truth]} \{‘Brooklyn Joseph Beckham’, ‘Romeo James Beckham’, ‘Cruz David Beckham’, ‘Harper Seven Beckham’\}
\end{reasonbox}
\end{table}

\begin{table}[H]
\centering
\caption{Example \framework{} response in \dataset{CWQ}. \textcolor{blue}{Blue} denotes responses generated by the \framework{} agent.}
\begin{reasonbox}[title=\framework{} Response Example - \dataset{CWQ}]
Question: What movie, written by Mike Jefferies, starred Nick Cannon? \\
Initial entities: "Mike Jefferies", "Nick Cannon"
\vspace{0.5em}\hrule\vspace{0.5em}
\begin{blueblock}
\texttt{<think>} \\
Let's start by finding a movie written by Mike Jefferies. We can use the relation that associates writers with their movies. \\
\texttt{</think>} \\
\texttt{<kg-query>} \\
$\gettailrelations$(\text{"Mike Jefferies"})
\texttt{</kg-query>} 
\end{blueblock}
\vspace{0.5em}\hrule\vspace{0.5em}
\texttt{<information>}Tail relations for entity "Mike Jefferies": \\
film.film\_story\_contributor.film\_story\_credits, people.person.nationality, people.person.place\_of\_birth, film.writer.film,  ... \texttt{</information>}
\vspace{0.5em}\hrule\vspace{0.5em}
\begin{blueblock}
\texttt{<think>} \\
From the initial query results, we have two possible options for Mike Jefferies' work. Let's start with "film.writer.film" to find the movies written by Mike Jefferies and then see if Nick Cannon starred in any of those films. \\
\texttt{</think>} \\
\texttt{<kg-query>} \\
$\gettailentities$("Mike Jefferies", "film.writer.film") \\
\texttt{</kg-query>}
\end{blueblock}
\vspace{0.5em}\hrule\vspace{0.5em}
\texttt{<information>}Tail entities for relation "film.writer.film" with head "Mike Jefferies": Goal II: Living the Dream, Goal III: Taking on the World\texttt{</information>} 
\vspace{0.5em}\hrule\vspace{0.5em}
\begin{blueblock}
\texttt{<think>} \\
Now that we have two movies written by Mike Jefferies, let's check if Nick Cannon has starred in any of these movies. We'll use the relation that connects actors with the movies they star in. \\
\texttt{</think>}
\texttt{<kg-query>} \\
$\getheadrelations$("Nick Cannon") \\
\texttt{</kg-query>} 
\end{blueblock}
\vspace{0.5em}\hrule\vspace{0.5em}
\texttt{<information>}Head relations for entity "Nick Cannon": 
base.popstra.dated.participant, film.personal\_film\_appearance.person, music.album.primary\_release, \\base.popstra.lived\_with.participant, film.film.written\_by, ... \texttt{</information>}
\vspace{0.5em}\hrule\vspace{0.5em}
...
\\
...
\\
...
\vspace{0.5em}\hrule\vspace{0.5em}
\begin{blueblock}
\texttt{<think>} \\
From the relations where "Goal II: Living the Dream" is involved, we can see that it is a film with a writer, which aligns with Mike Jefferies. \\
\texttt{</think>} \\
\texttt{<answer>} \\
Goal II: Living the Dream, School Dance, School Gyrls, Goal III: Taking on the World \\
\texttt{</answer>}
\end{blueblock}
\vspace{0.5em}\hrule\vspace{0.5em}
\vspace{0.1em}\hrule\vspace{0.5em}

\textbf{[ground\_truth]} \{'Goal II: Living the Dream' \}
\end{reasonbox}
\end{table}

\begin{table}[H]
\centering
\caption{Example \framework{} response in \dataset{SimpleQA}. \textcolor{blue}{Blue} denotes responses generated by the \framework{} agent.}
\begin{reasonbox}[title=\framework{} Response Example - \dataset{SimpleQA}]
Question: Where is the madam satan located? (Initial entities: "Madam Satan") \\
\vspace{0.5em}\hrule\vspace{0.5em}
\begin{blueblock}
\texttt{<think>} \\
I need to find out where Madam Satan is located. First, I will check if it has a location or if it's an entity that can be queried about its location. \\
\texttt{</think>} \\
\texttt{<kg-query>} \\
$\gettailrelations(\text{"Madam Satan"})$ \\
\texttt{</kg-query>}
\end{blueblock}

\vspace{0.5em}\hrule\vspace{0.5em}

\texttt{<information>}Tail relations for entity "Madam Satan": \\
film.film.cinematography, common.topic.notable\_types, film.film.language, 
film.film.starring,film.film.country,   ... award.ranked\_item.appears\_in\_ranked\_lists,  
film.film.film\_art\_direction\_by\texttt{</information>}

\vspace{0.5em}\hrule\vspace{0.5em}

\begin{blueblock}
\texttt{<think>} \\
None of the identified relations are directly related to location. Let me check if Madam Satan is detailed in a geographical context. \\
\texttt{</think>} \\
\texttt{<kg-query>} \\
$\gettailentities(\text{"Madam Satan"}, \text{"film.film.country"})$ \\
\texttt{</kg-query>}
\end{blueblock}

\vspace{0.5em}\hrule\vspace{0.5em}

\texttt{<information>}Tail entities for relation "film.film.country" with head "Madam Satan": United States of America\texttt{</information>}

\vspace{0.5em}\hrule\vspace{0.5em}

\begin{blueblock}
\texttt{<think>} \\
I have found that Madam Satan, identified as a film, is located in the United States of America. However, it's crucial to confirm the specific location or city within the country where Madam Satan was filmed. \\
\texttt{</think>} \\
\texttt{<kg-query>} \\
$\gettailentities(\text{"Madam Satan"}, \text{"film.film.starring"})$ \\
\texttt{</kg-query>}
\end{blueblock}

\vspace{0.5em}\hrule\vspace{0.5em}
..\\
..\\
..
\vspace{0.5em}\hrule\vspace{0.5em}

\texttt{<information>}Tail entities for relation "film.film.runtime" with head "Madam Satan": m.03kx9f2\texttt{</information>}

\vspace{0.5em}\hrule\vspace{0.5em}

\begin{blueblock}
\texttt{<think>} \\
I am still not obtaining the most helpful information about the location. Possibly Madam Satan was released, but not filmed in a specific place. \\
\texttt{</think>} \\
\texttt{<kg-query>} \\
$\getheadrelations(\text{"m.02tb01n"})$ \\
\texttt{</kg-query>}
\end{blueblock}

\vspace{0.5em}\hrule\vspace{0.5em}

\texttt{<information>}Head relations for entity "m.02tb01n": \\
film.film.starring, film.actor.film\texttt{</information>}

\vspace{0.5em}\hrule\vspace{0.5em}

\begin{blueblock}
\texttt{<think>} \\
I have questioned the relation "film.actor.film". This relation includes the movies the actor is in. Let's try to find a clearer detail for the location of Madam Satan. \\
\texttt{</think>} \\
\texttt{<answer>} \\
United States of America \\
\texttt{</answer>}
\end{blueblock}
\vspace{0.5em}\hrule\vspace{0.5em}
\vspace{0.1em}\hrule\vspace{0.5em}

\textbf{[ground\_truth]} \{'United States of America' \}

\end{reasonbox}
\end{table}

\begin{table}[H]
\centering
\caption{Example \framework{} response in \dataset{T-REx}. \textcolor{blue}{Blue} denotes responses generated by the \framework{} agent.}
\begin{reasonbox}[title=\framework{} Response Example — \dataset{T-REx}]
Question: What is the instance of Solberg?? \\
Initial entities: "Solberg"
\vspace{0.5em}\hrule\vspace{0.5em}

\begin{blueblock}
\texttt{<think>} \\
Let's start with the entity Solberg and try to find out what it represents. It seems like Solberg could be an entity that describes a geopolitical boundary, location, or characterization. The query can be adjusted from general to specific as needed. \\
\texttt{</think>} \\
\texttt{<kg-query>} \\
$\gettailrelations(\text{"Solberg"})$ \\
\texttt{</kg-query>}
\end{blueblock}

\vspace{0.5em}\hrule\vspace{0.5em}
\texttt{<information>}Tail relations for entity "Solberg": P1889, P407, P1552, P282, P31, P460\texttt{</information>}
\vspace{0.5em}\hrule\vspace{0.5em}

\begin{blueblock}
\texttt{<think>} \\
The relations provided show how Solberg is related to other entities. Let's explore each relation to find out more about Solberg. \\
\texttt{</think>} \\
\texttt{<kg-query>} \\
$\gettailentities(\text{"Solberg"},\ \text{"P282"})$ \\
\texttt{</kg-query>}
\end{blueblock}

\vspace{0.5em}\hrule\vspace{0.5em}

\texttt{<information>}Tail entities for relation "P282" with head "Solberg": Latin script\texttt{</information>}

\vspace{0.5em}\hrule\vspace{0.5em}

\begin{blueblock}
\texttt{<think>} \\
The relation P282 corresponds to "type.object.name" according to the API documentation, indicating that Solberg is a human name. The relation P407 is also noteworthy but does not give further specific information. Let's try a broader query for more context related to Solberg. \\
\texttt{</think>} \\
\texttt{<kg-query>} \\
$\gettailentities(\text{"Solberg"},\ \text{"P31"})$ \\
\texttt{</kg-query>}
\end{blueblock}
\vspace{0.5em}\hrule\vspace{0.5em}
\texttt{<information>}Tail entities for relation "P31" with head "Solberg": Wikimedia disambiguation page, family name\texttt{</information>}
\vspace{0.5em}\hrule\vspace{0.5em}
...\\
...\\
...
\vspace{0.5em}\hrule\vspace{0.5em}

\begin{blueblock}
\texttt{<think>} \\
The retrieval of "Soelberg", "Sohlberg", and "Sohlberg" suggests that Solberg might be associated with these names. To find an instance of Solberg, we can search for a specific entity of this name. \\
\texttt{</think>} \\
\texttt{<answer>} \\
Latin script, Wikimedia disambiguation page, family name, Soelberg, Sohlberg, Sohlberg \\
\texttt{</answer>}

\end{blueblock}
\vspace{0.5em}\hrule\vspace{0.5em}
\vspace{0.1em}\hrule\vspace{0.5em}

\textbf{[ground\_truth]} \{'family name', 'human settlement', 'urban area in Sweden', 'minor locality in Sweden', 'minor locality in Sweden', 'Wikimedia disambiguation page', 'human settlement', 'Ortsteil', 'stone setting', 'human settlement', 'minor locality in Sweden', 'hill', 'stone setting', 'hill', 'human settlement' \}

\end{reasonbox}
\end{table}

\begin{table}[H]
\centering
\caption{Example \framework{} response in \dataset{QALD10en}. \textcolor{blue}{Blue} denotes responses generated by the \framework{} agent.}
\begin{reasonbox}[title=\framework{} Response Example - \dataset{QALD10en}]
Question: After whom is the Riemannian geometry named?? (Initial entities: "Riemannian geometry") \\
\vspace{0.5em}\hrule\vspace{0.5em}

\begin{blueblock}
\texttt{<think>} \\
I need to find out who the Riemannian geometry is named after. Let's start by investigating the relations and entities connected to "Riemannian geometry". \\
\texttt{</think>} \\
\texttt{<kg-query>} \\
$\gettailrelations(\text{"Riemannian geometry"})$ \\
\texttt{</kg-query>}
\end{blueblock}

\vspace{0.5em}\hrule\vspace{0.5em}

\texttt{<information>}Tail relations for entity "Riemannian geometry": \\
on focus list of Wikimedia project, named after, maintained by WikiProject, topic's main category, instance of, subclass of\texttt{</information>}

\vspace{0.5em}\hrule\vspace{0.5em}

\begin{blueblock}
\texttt{<think>} \\
From the relations extracted, it appears the Riemannian geometry is named after someone. I will now search for the person it's named after. \\
\texttt{</think>} \\
\texttt{<kg-query>} \\
$\gettailentities(\text{"Riemannian geometry"}, \text{"named after"})$ \\
\texttt{</kg-query>}
\end{blueblock}

\vspace{0.5em}\hrule\vspace{0.5em}

\texttt{<information>}Tail entities for relation "named after" with head "Riemannian geometry": Bernhard Riemann\texttt{</information>}

\vspace{0.5em}\hrule\vspace{0.5em}

\begin{blueblock}
\texttt{<think>} \\
I now have the information that the Riemannian geometry is named after Bernhard Riemann. Let's confirm this by checking if there's more detailed information about this relationship. \\
\texttt{</think>} \\
\texttt{<kg-query>} \\
$\getheadrelations(\text{"Bernhard Riemann"})$ \\
\texttt{</kg-query>}
\end{blueblock}

\vspace{0.5em}\hrule\vspace{0.5em}

\texttt{<information>}Head relations for entity "Bernhard Riemann": \\
named after\texttt{</information>}

\vspace{0.5em}\hrule\vspace{0.5em}

\begin{blueblock}
\texttt{<think>} \\
These results are confirming that Bernhard Riemann is indeed the person the Riemannian geometry is named after. No further information is needed. \\
\texttt{</think>} \\
\texttt{<kg-query>} \\
$\getheadentities(\text{"Bernhard Riemann"}, \text{"named after"})$ \\
\texttt{</kg-query>}
\end{blueblock}

\vspace{0.5em}\hrule\vspace{0.5em}

\texttt{<information>}Head entities for relation "named after" with tail "Bernhard Riemann": Riemannian geometry\texttt{</information>}

\vspace{0.5em}\hrule\vspace{0.5em}

\begin{blueblock}
\texttt{<think>} \\
This confirms my findings; the Riemannian geometry is named after Bernhard Riemann. \\
\texttt{</think>} \\
\texttt{<answer>} \\
Bernhard Riemann \\
\texttt{</answer>}
\end{blueblock}

\vspace{0.5em}\hrule\vspace{0.5em}
\vspace{0.1em}\hrule\vspace{0.5em}

\textbf{[ground\_truth]} \{'Bernhard Riemann'\}

\end{reasonbox}
\end{table}

\begin{table}[H]
\centering
\caption{Example \framework{} response in \dataset{GrailQA}. \textcolor{blue}{Blue} denotes responses generated by the \framework{} agent.}
\begin{reasonbox}[title=\framework{} Response Example — \dataset{GrailQA}]
Question: What video game is published by softdisk? \\
Initial entities: "Near West Side"
\vspace{0.5em}\hrule\vspace{0.5em}

\begin{blueblock}
\texttt{<think>} \\
First, I need to find out what game is published by Softdisk. Without specific knowledge about "Near West Side", I will start with the game publisher for Softdisk. \\
\texttt{</think>} \\
\texttt{<kg-query>} \\
$\getheadrelations(\text{"Near West Side"})$ \\
\texttt{</kg-query>}
\end{blueblock}

\vspace{0.5em}\hrule\vspace{0.5em}

\texttt{<information>}Head relations for entity "Near West Side": \\
book.periodical\_publisher\_period.publisher, business.industry.companies, business.employment\_tenure.company\texttt{</information>}

\vspace{0.5em}\hrule\vspace{0.5em}

\begin{blueblock}
\texttt{<think>} \\
The head relation "business.employment\_tenure.company" suggests that "Near West Side" might be a game company or related to a game. I will try to find out the publisher of a game related to "Near West Side" by looking for games published by Softdisk. \\
\texttt{</think>} \\
\texttt{<kg-query>} \\
$\gettailrelations(\text{"Near West Side"})$ \\
\texttt{</kg-query>}
\end{blueblock}

\vspace{0.5em}\hrule\vspace{0.5em}

\texttt{<information>}Tail relations for entity "Near West Side": \\
organization.organization.place\_founded, organization.organization.headquarters, book.periodical\_publisher.periodicals\_published, business.employer.employees, cvg.cvg\_publisher.games\_published, business.business\_operation.industry, common.topic.notable\_types\texttt{</information>}

\vspace{0.5em}\hrule\vspace{0.5em}
...\\
...\\
...\\
\vspace{0.5em}\hrule\vspace{0.5em}

\begin{blueblock}
\texttt{<think>} \\
The previous query did not provide a conclusive answer. Let's try a different approach by checking if "Augusto Cury" is associated with any game publishers. \\
\texttt{</think>} \\
\texttt{<answer>} \\
Augusto Cury, Catacomb 3-D, Chagunitzu, Dangerous Dave in the Haunted Mansion, Hovertank 3D, Iron Seed, Lado Skorin, North Cambridge Academy, Roberto Torretti, Shadow Knights, Star Film Company, Tiles of the Dragon \\
\texttt{</answer>}
\end{blueblock}

\vspace{0.5em}\hrule\vspace{0.5em}
\vspace{0.1em}\hrule\vspace{0.5em}

\textbf{[ground\_truth]} \{'Rescue Rover','Catacomb 3-D','Shadow Knights',"Dangerous Dave's Risky Rescue",'Tiles of the Dragon', 'Catacomb', 'Commander Keen: Keen Dreams', 'Dave Goes Nutz!','Chagunitzu','Dangerous Dave in the Haunted Mansion','Hovertank 3D','Iron Seed'\}

\end{reasonbox}
\end{table}

\newpage
\section*{NeurIPS Paper Checklist}

\begin{enumerate}

\item {\bf Claims}
    \item[] Question: Do the main claims made in the abstract and introduction accurately reflect the paper's contributions and scope?
    \item[] Answer: \answerYes{} %
    \item[] Justification: The abstract and introduction clearly state the paper’s main claims: improving KG-RAG effectiveness, efficiency, and transferability through reinforcement learning over structured KG interactions. These claims are supported by experimental results in \Cref{sec:exp1}--\ref{sec:exp3}.

\item {\bf Limitations}
    \item[] Question: Does the paper discuss the limitations of the work performed by the authors?
    \item[] Answer: \answerYes{} %
    \item[] Justification: We discuss the limitations of \framework{} in Appendix~\ref{sec:discussion}, including its dependence on knowledge graph quality, potential degradation under schema mismatch, and the computational overhead of reinforcement learning training.

\item {\bf Theory assumptions and proofs}
    \item[] Question: For each theoretical result, does the paper provide the full set of assumptions and a complete (and correct) proof?
    \item[] Answer: \answerYes{} %
    \item[] Justification: Yes. We provide formal propositions and proofs for the completeness, finite-horizon realizability, schema-free transferability, and minimality of the KG action interface in Appendix~\ref{appendix:theoretical_proof}. The assumptions and operator definitions are explicitly stated in the preliminaries.

    \item {\bf Experimental result reproducibility}
    \item[] Question: Does the paper fully disclose all the information needed to reproduce the main experimental results of the paper to the extent that it affects the main claims and/or conclusions of the paper (regardless of whether the code and data are provided or not)?
    \item[] Answer: \answerYes{} %
    \item[] Justification: We provide full methodological details, including action design, reward formulation, training procedures, hyperparameters, and evaluation protocols in the main paper and appendix. Code is released anonymously for reproducibility.

\item {\bf Open access to data and code}
    \item[] Question: Does the paper provide open access to the data and code, with sufficient instructions to faithfully reproduce the main experimental results, as described in supplemental material?
    \item[] Answer: \answerYes{} %
    \item[] Justification: We release anonymized code, training scripts, and evaluation pipelines (link provided in the abstract). All benchmark datasets used in this work are publicly available.

\item {\bf Experimental setting/details}
    \item[] Question: Does the paper specify all the training and test details (e.g., data splits, hyperparameters, how they were chosen, type of optimizer) necessary to understand the results?
    \item[] Answer: \answerYes{} %
    \item[] Justification: We specify the benchmark datasets, model backbones, training horizon, reward settings, RL optimization configurations, and evaluation metrics in \Cref{sec:experiments} and the appendix.

\item {\bf Experiment statistical significance}
    \item[] Question: Does the paper report error bars suitably and correctly defined or other appropriate information about the statistical significance of the experiments?
    \item[] Answer: \answerYes{} %
    \item[] Justification:  We report mean and standard deviation across five runs in all main experimental tables and ablation studies. The reported error bars reflect run-to-run variation.

\item {\bf Experiments compute resources}
    \item[] Question: For each experiment, does the paper provide sufficient information on the computer resources (type of compute workers, memory, time of execution) needed to reproduce the experiments?
    \item[] Answer: \answerYes{} %
    \item[] Justification: We provide hardware specifications, memory configurations, precision settings, and training duration details in the appendix, including GPU type and model-specific resource allocation.
    
\item {\bf Code of ethics}
    \item[] Question: Does the research conducted in the paper conform, in every respect, with the NeurIPS Code of Ethics \url{https://neurips.cc/public/EthicsGuidelines}?
    \item[] Answer: \answerYes{} %
    \item[] Justification: We have reviewed the NeurIPS Code of Ethics and confirm that this work complies with its guidelines. The experiments use publicly available datasets and models without involving human subjects or sensitive personal data.

\item {\bf Broader impacts}
    \item[] Question: Does the paper discuss both potential positive societal impacts and negative societal impacts of the work performed?
    \item[] Answer: \answerYes{} %
    \item[] Justification: We discuss both positive societal impacts and potential negative impacts in Appendix~\ref{sec:discussion}, including improved factual reasoning and risks of propagating incorrect or manipulated knowledge.
    
\item {\bf Safeguards}
    \item[] Question: Does the paper describe safeguards that have been put in place for responsible release of data or models that have a high risk for misuse (e.g., pre-trained language models, image generators, or scraped datasets)?
    \item[] Answer: \answerNA{} %
    \item[] Justification: This work does not release high-risk models or sensitive datasets. The proposed method builds upon existing open-weight language models and public benchmark datasets.

\item {\bf Licenses for existing assets}
    \item[] Question: Are the creators or original owners of assets (e.g., code, data, models), used in the paper, properly credited and are the license and terms of use explicitly mentioned and properly respected?
    \item[] Answer: \answerYes{} %
    \item[] Justification: All datasets, models, and baseline implementations used in this work are properly cited, and their original licenses and usage terms are respected.

\item {\bf New assets}
    \item[] Question: Are new assets introduced in the paper well documented and is the documentation provided alongside the assets?
    \item[] Answer: \answerYes{} %
    \item[] Justification: We release the implementation of \framework{} with documentation and scripts for reproducing the main experimental results.

\item {\bf Crowdsourcing and research with human subjects}
    \item[] Question: For crowdsourcing experiments and research with human subjects, does the paper include the full text of instructions given to participants and screenshots, if applicable, as well as details about compensation (if any)? 
    \item[] Answer: \answerNA{} %
    \item[] Justification: This work does not involve crowdsourcing or research with human subjects.

\item {\bf Institutional review board (IRB) approvals or equivalent for research with human subjects}
    \item[] Question: Does the paper describe potential risks incurred by study participants, whether such risks were disclosed to the subjects, and whether Institutional Review Board (IRB) approvals (or an equivalent approval/review based on the requirements of your country or institution) were obtained?
    \item[] Answer: \answerNA{} %
    \item[] Justification: This work does not involve human subjects and does not require IRB approval.

\item {\bf Declaration of LLM usage}
    \item[] Question: Does the paper describe the usage of LLMs if it is an important, original, or non-standard component of the core methods in this research? Note that if the LLM is used only for writing, editing, or formatting purposes and does \emph{not} impact the core methodology, scientific rigor, or originality of the research, declaration is not required.
    \item[] Answer: \answerYes{} %
    \item[] Justification: Large language models are a core component of the proposed KG-R1 framework and are fully described in the methodology section. Additionally, we used an LLM-based coding assistant to support parts of the implementation. All generated code was reviewed and validated by the authors, and all experiments were designed, executed, and verified by the authors.

\end{enumerate}

\end{document}